\documentclass[onecolumn]{article}

\usepackage{arxiv}

\usepackage[utf8]{inputenc} 
\usepackage[T1]{fontenc}    
\usepackage{hyperref}       
\usepackage{url}            
\usepackage{booktabs}       
\usepackage{amsfonts}       
\usepackage{nicefrac}       
\usepackage{microtype}      
\usepackage{lipsum}
\usepackage{graphicx}
\graphicspath{ {./images/} }

\usepackage{algorithm}
\usepackage{algorithmic}
\usepackage{graphicx}
\usepackage{amsmath}
\usepackage{amsthm}
\usepackage{booktabs}
\usepackage{color}
\usepackage{bm}
\usepackage{amssymb}
\usepackage{float}
\usepackage{subfig}
\usepackage{enumitem}
\usepackage{wrapfig} 
\usepackage{multirow}
\usepackage{lipsum}

\newtheorem{assumption}{Assumption}
\newtheorem{proposition}{Proposition}
\newtheorem{lemma}{Lemma}

\newtheorem{theorem}{Theorem}
\newtheorem{definition}{Definition}
\newtheorem{remark}{Remark}

\usepackage{setspace}

\DeclareMathOperator{\E}{\mathbb{E}}

\DeclareMathOperator{\dist}{dist}

\DeclareMathOperator{\sign}{sign}

\newcommand{\ip}[2]{\left\langle#1,#2\right\rangle}
\newcommand{\norm}[1]{\left\lVert#1\right\rVert}

\renewcommand{\[}{\left[}
\renewcommand{\]}{\right]}

\def \R {\mathbb{R}}

\def \S {\mathbb{S}}

\def \b {\beta}

\def \< {\langle}
\def \> {\rangle}

\def \vd {\bm{d}}

\def \vg {\bm{g}}
\def \vh {\bm{h}}

\def \vu {\bm{u}}

\def \vw {\bm{w}}
\def \vy {\bm{y}}

\def \vX {\bm{X}}

\def \vI {\bm{I}}

\title{Sparse Personalized Federated Learning}

\author{Xiaofeng Liu$^*$, Yinchuan Li$^*$, Qing Wang, Xu Zhang, Yunfeng Shao, Yanhui Geng
}

\begin{document}

\date{}
\maketitle
\newcommand\blfootnote[1]{
\begingroup
\renewcommand\thefootnote{}\footnote{#1}
\addtocounter{footnote}{-1}
\endgroup
}

\blfootnote{Xiaofeng Liu, Qing Wang are with School of Electrical and Information Engineering, Tianjin University, Tianjin, China (e-mail: xiaofengliull@tju.edu.cn, wangq@tju.edu.cn)

Yinchuan Li, Yunfeng Shao, Yanhui Geng are with Huawei Noah's Ark Lab, Beijing, China (e-mail: liyinchuan@huawei.com, shaoyunfeng@huawei.com, geng.yanhui@huawei.com)

Xu Zhang is with LSEC, Academy of Mathematics and Systems Science, Chinese Academy of Sciences, Beijing, China (e-mail: xuzhang\_cas@lsec.cc.ac.cn)

$^*$ Equal Contribution. This work was completed while Xiaofeng Liu was a member of the Huawei Noah's Ark Lab for Advanced Study.

Corresponding author: Xu Zhang, Qing Wang.
}

\begin{abstract}
Federated Learning (FL) is a collaborative machine learning technique to train a global model without obtaining clients' private data. The main challenges in FL are statistical diversity among clients, limited computing capability among clients' equipments, and the excessive communication overhead between the server and clients. To address these challenges, we propose a novel sparse personalized federated learning scheme via maximizing correlation ({\texttt{FedMac}}).  By incorporating  an approximated $\ell_1$-norm and the correlation between client models and global model into standard FL loss function, the performance on statistical diversity data is improved and the communicational and computational loads required in the network are reduced compared with non-sparse FL. Convergence analysis shows that the sparse constraints in {\texttt{FedMac}} do not affect the convergence rate of the global model, and theoretical results show that {\texttt{FedMac}} can achieve good sparse personalization, which is better than the personalized methods based on $\ell_2$-norm.  Experimentally, we demonstrate the benefits of this sparse personalization architecture compared with the state-of-the-art personalization methods (e.g. {\texttt{FedMac}} respectively achieves 98.95\%, 99.37\%, 90.90\%, 89.06\% and \textcolor{black}{73.52\%} accuracy on the MNIST, FMNIST, CIFAR-100, Synthetic and \textcolor{black}{CINIC-10} datasets under non-i.i.d. variants).
\end{abstract}



\section{Introduction}

Federated learning (FL)~\cite{mcmahan2017communication,li2020federated, 8889996} has grown rapidly in recent years thanks to the popularity of hand-held devices such as mobile phones and tablets~\cite{mothukuri2021survey,rieke2020future,sheller2020federated,li2022avoid}. Since the training data are usually distributed on clients and are mostly private, FL needs to establish a network of clients connected to a server and train a global model in a privacy-preserving manner. One of the main challenge of FL is the statistical diversity, i.e., the data distributions among clients are distinct (i.e., non-i.i.d.). Recently, personalized FL~\cite{fallah2020personalized,t2020personalized,li2022federated,zhang2022personalized} is proposed to address this problem, which motivates us to develop the approach to achieve better personalization. Another challenge is that the server and clients need multiple rounds of communication to guarantee the convergence performance. However, the communication latency from clients to the cloud server is high and the communication bandwidth is limited.  Hence, communication-efficient approaches have to be proposed~\cite{8889996, 8945292, 9305988, zong2021communication}.

Many federated learning algorithms~\cite{mcmahan2017communication,t2020personalized,li2018federated} need to equalize the dimensions of the client models and the global model. However, the global model generally requires more parameters than the client models to cover the main features of all clients, which means there might be too many parameters in client models.  As a result, it is possible to prune many parameters from client networks without affecting performance. In this way, parameter traffic between the client models and the global model can be directly reduced, and the required storage and computing amount are reduced~\cite{he2019bag}, which is crucial for applications running on low-capacity hand-held devices. Unfortunately, traditional sparse optimizers~\cite{srinivas2017training,han2015learning,gale2019state,li2021structured} cannot obtain personalized models in client networks, which results in poor performance and convergence rate on statistical diversity data~\cite{t2020personalized}. 

Inspired by broad applications of personalized models in healthcare, finance and AI services~\cite{deng2020adaptive}, this paper proposes a sparse personalized federated learning scheme via maximizing correlation ({\texttt{FedMac}}). By incorporating an approximated $\ell_1$-norm and the correlation (inner product) of global model and local models into the loss function, the proposed method achieves better personalization ability and higher communication efficiency. This reason is as follows: the introduction of correlation makes each client leverage global model to optimize its personalized model w.r.t. its own data and so improves the performance on statistical diversity data; the incorporation of approximated $\ell_1$-norm makes the models become sparse and so the amount of communication is greatly reduced.

\subsection{Main Contributions}

In this paper, we propose a novel sparse personalized FL method based on maximizing correlation, and further formulate an optimization problem designed for {\texttt{FedMac}} by using $\ell_1$-norms and the correlation between the global model and client models in a regularized loss function. The $\ell_1$-norm constraints can generate sparse models, and hence the communication loads between the server and the clients are greatly reduced, i.e., there are many zero weights in models and only the non-zero weights need to be uploaded and downloaded. Maximizing the correlation between the global model and client models encourages clients to pursue their own models with different directions, but not to stay far away from the global model.

In addition, we propose an approximate algorithm to solve {\texttt{FedMac}}, and present convergence analysis for the proposed algorithm, which indicates that the sparse constraints do not affect the convergence rate of the global model. Furthermore, we theoretically prove that the performance of {\texttt{FedMac}} under sparse conditions is better than that of the personalization methods based on $\ell_2$-norm~\cite{t2020personalized,li2018federated,mcmahan2017communication}. The training data required for the proposed {\texttt{FedMac}} is significantly less than that required for the methods based on $\ell_2$-norm.

Finally, we evaluate the performance of {\texttt{FedMac}} using real datasets that capture the statistical diversity of clients’ data. Experimental results show that {\texttt{FedMac}} is superior to various advanced personalization algorithms, and our algorithm can make the network sparse to reduce the amount of communication. Experimental results based on different deep neural networks show that our {\texttt{FedMac}} achieves the highest accuracy on both the personalized model (98.95\%, 99.37\%, 90.90\%, 89.06\%, and \textcolor{black}{73.52\%}) and the global model (96.90\%, 85.80\%, 86.57\%, 84.89\%, and \textcolor{black}{68.97\%})  on the MNIST, FMNIST, CIFAR-100, Synthetic and \textcolor{black}{CINIC-10} datasets, respectively.

{\color{black}
\subsection{Organization and Main Notations}

The remainder of the paper is organized as follows. 
Section~\ref{sec_bk} reviews the basic FL framework and the major related works.
In Section~\ref{sec_t_sp}, we explore the potential better personalization constraint.
The problem formulation and algorithm are formulated in Section~\ref{sec_FedMac}. 
Section~\ref{sec_theoretical_analysis} shows the convergence analysis with some necessary Lemmas and Theorems. 
Section~\ref{sec_experiment_results} presents the experimental results, followed by some analysis. 
Finally, conclusion and discussions are given in Section~\ref{sec_conclusion}.

The main notations used are listed below.
}
\begin{table}[htpb]
  \centering
  {
  \color{black}
  \scalebox{1}{
  \begin{tabular}{lc}
    \toprule
    Definition
    & Notation
     \\
    \midrule
    Number of clients & $N$ \\ 
    Global model & $\bm w \in \mathbb{R}^{d}$ \\ 
    Personalized model of the $i$-th client & $\bm \theta_i \in \mathbb{R}^{d}$ \\
    Optimal global model & $\bm w^{\star}$ \\
    Local global model & $\bm w_i$ \\
    Optimal personalized model  & $\bm \theta^{\star}_i$ \\
    Loss function & $\mathcal{L}(\cdot)$ \\
    Training data of $i$-th client & $Z_{i}$ \\
    Expected loss of $i$-th client & $\ell_{i} \triangleq \mathbb{E}_{{\mathcal Z}_i} \mathcal{L} $ \\
    Hyperparameter for sparsity & $\gamma$ \\
    Hyperparameter for personalization & $\lambda$ \\
    Hyperparameter for aggregation & $\beta$ \\
    Hyperparameter $\zeta$ & $\zeta \triangleq {\lambda}/{ \gamma }$ \\
    Global training round & $T$ \\
    Local training round & $R$ \\
    Number of clients to aggregate & $S$ \\
    Learning rate for updating $\bm w$ & $\eta$ \\
    Learning rate for updating $\bm \theta_i$ & $\eta_p$ \\
    \bottomrule
  \end{tabular}}
}
\end{table}

{
\color{black}
\section{Background and Related work}\label{sec_bk}


\subsection{Conventional FL Methods}
Suppose there are $N$ clients communicating with a server, a federated learning system aims to find a {\em global model} $\bm w \in \mathbb{R}^{d}$ by minimizing
\begin{align}
\label{conv-FL}
    \min _{\bm w \in \mathbb{R}^{d}}\left\{\ell(\bm w)\triangleq \frac{1}{N} \sum_{i=1}^{N} \ell_{i}(\bm w)\right\},
\end{align}
where $\ell_{i}(\cdot): \mathbb{R}^{d} \rightarrow \mathbb{R},~ i=1, \ldots, N$ is the expected loss over the data distribution of the $i$-th client. Let $Z_{i}=\left\{(X_i,Y_i)\right\}$ denote the training data randomly drawn from the distribution of client $i$, where $X_i$ is the input variable while $Y_i$ is the response variable. Note that clients might have non-i.i.d. data distributions, i.e., the distributions of $Z_{i}$ and $Z_j\,(i \neq j)$ might be distinct, since their data may come from different environments, contexts and applications~\cite{t2020personalized}. Let $h(x;\bm w)$ be the output of an $L$-layer feedforward overparameterized deep neural network, and $\mathcal{L}(h;y)$ be the loss function, then we have 
\begin{align}
    \ell_{i}(\bm w)\triangleq \mathbb{E}_{{\mathcal Z}_i}\left[      \mathcal{L}(h(X_i;\bm w);Y_i)     \right],
\end{align}
where $\mathcal{Z}_i$ is the distributions of $Z_{i}$.

\subsection{Personalized FL}
FedAvg~\cite{mcmahan2017communication} is known as the first FL algorithm to build a global model from a subset of clients with decentralized data, where the client models are updated by the local SGD method. However, it cannot handle the problem of non-i.i.d. data distribution well~\cite{zhu2021federated}. To address this problem, various personalized FL methods have been proposed, which can be divided into different categories mainly including local fine-tuning based methods~\cite{fallah2020personalized, huang2021personalized, li2018federated, t2020personalized}, personalization layer based methods~\cite{arivazhagan2019federated}, multi-task learning based methods~\cite{smith2017federated}, knowledge distillation based methods~\cite{hitaj2017deep}  and lifelong learning based methods~\cite{liu2019lifelong}. The methods based on the personalization layer require the client to permanently store its personalization layer without releasing it. And the methods based on multi-task learning, knowledge distillation, and lifelong learning introduce other complex mechanisms to achieve personalization in FL, which may encounter difficulties in computation, deployment, and generalization. Therefore, in this paper, we pay more attention to personalized methods based on local fine-tuning.

\subsection{Local fine-tuning based personalized FL}  
Local fine-tuning is the most classic and powerful personalization method~\cite{zhu2021federated}. It aims to fine-tune the local model through meta-learning, regularization, interpolation and other techniques to combine local and global information. Per-FedAvg~\cite{fallah2020personalized} sets up an initial meta-model that can be updated effectively after one more gradient descent step.
By facilitating pairwise collaborations between clients with similar data, FedAMP~\cite{huang2021personalized}  uses federated attentive message passing to promote similar clients to collaborate more. The
$\ell_2$-norm regularization based methods such as Fedprox~\cite{li2018federated} and pFedMe~\cite{t2020personalized}  also achieve good personalization. However, none of these works are specifically designed for the communication-efficient and computation-friendly FL framework, which prompts us to propose \texttt{FedMac}, a sparse personalized FL method.
}

\begin{figure*}[htbp]
	\centering
	\subfloat{
	\includegraphics[height=1.6in]{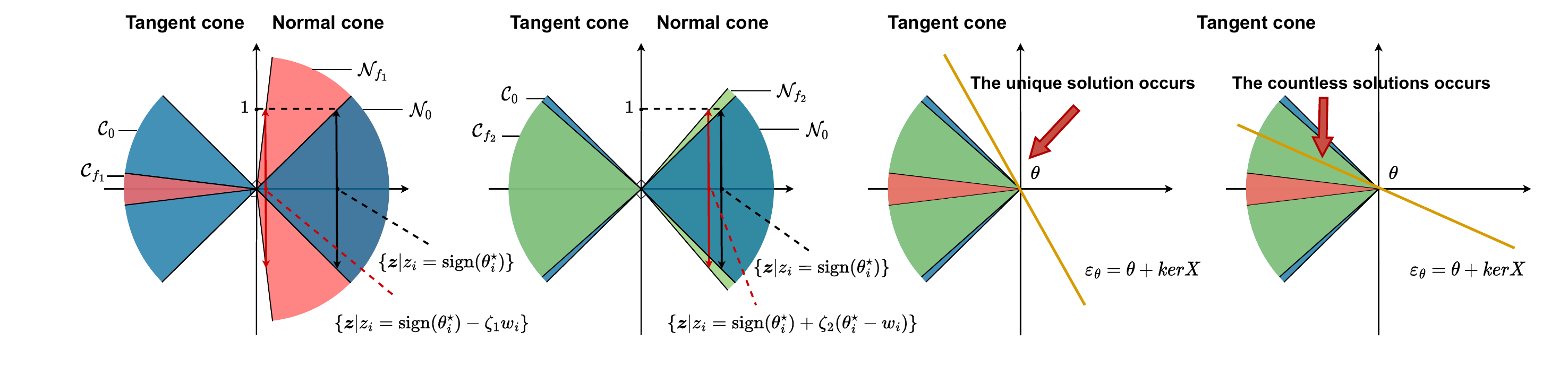}
	}
\caption{\color{black}The changes of normal cone and tangent cone under different personalized constraints with $\bm \theta^\star=[1,0]$. \textbf{Left:} Maximizing correlation constraint $f_1(\bm \theta) = \| \bm \theta \|_1 - {\zeta_1} \langle  \bm \theta , {\bm w} \rangle $; \textbf{Middle Left:} $\ell_2$-norm constraint $f_2(\bm \theta) = \| \bm \theta \|_1 + \frac{\zeta_2}{2} \|  \bm \theta - {\bm w} \|_2^2$;  \textbf{Middle Right:} The tangency occurs iff the affine subspace ${\mathcal{E}}_{\bm \theta}$ is disjoint from the spherical part of the tangent cone at point $\bm \theta$. 
\textbf{Right:} There are countless solutions iff the affine subspace ${\mathcal{E}}_{\bm \theta}$ intersects the spherical part of the tangent cone at point $\bm \theta$.
We draw the standard tangent cone $\mathcal{C}_0$ and normal cone $\mathcal{N}_0$ in black, draw the tangent cone $\mathcal{C}_{f_1}$ and normal cone $\mathcal{N}_{f_1}$ in red for $f_1(\bm \theta)$ and draw the tangent cone $\mathcal{C}_{f_2}$ and normal cone $\mathcal{N}_{f_2}$ in green for $f_2(\bm \theta)$. The global parameter $w$ is close to the optimal personalized parameter $\theta^\star$. It's known that the smaller the tangent cone, the higher the estimation precision for the same amount of training data. The figures show that $\mathcal{C}_{f_1}$ is much smaller than $\mathcal{C}_0$ while $\mathcal{C}_{f_2}$ is almost the same as $\mathcal{C}_0$, which means maximizing correlation constraint has higher estimation precision than $\ell_2$-norm constraint
under the same amount of training data.}

	\label{fig_superisorty}
\end{figure*}


\section{Towards Better Sparse Personalization}\label{sec_t_sp}

On the client side, we only consider one client at a time and omit the subscript $i$ for simplicity. To solve the statistical diversity problem and reduce the communication burden, we aim to find the {\em sparse personalized model} $\bm \theta$ by minimizing
\begin{align}
    \label{optimal-SP}
     \min _{\bm \theta \in \mathbb{R}^{d}} \left\{\ell \left(\bm \theta \right)    +  \gamma \| \bm \theta \|_1 + \lambda \bar f( \bm \theta, {\bm w})  \right\}, 
\end{align}
where $\gamma$ and $\lambda$ are weighting factors that respectively control the sparsity level and the degree of personalization, and $\bar f( \bm \theta, {\bm w})$ is a personalization constraint, allowing clients to benefit from the abundant data aggregation in the global model while maintaining a certain degree of personalization. In this section, our goal is to find a better personalization constraints under sparse conditions. We think that a good personalization constraint can enable clients to achieve better personalization capabilities, and can also improve the convergence speed of personalized model during the training process. In other words, even when the amount of client data is small, a high-precision personalized model can be trained with reference to the global model. 

To do this, we first simplify the problem to explore the potential better personalization constraint. Consider a linear neural network system, e.g., a single-layer network. Suppose that $\bm X \in \mathbb{R}^{N_D \times N_I}$ is the training input with $N_D$ being the number of training data and $N_I = d$ being the length of data, $\bm y \in \mathbb{R}^{N_D \times 1}$ is the training output (label) and $\bm \theta \in \mathbb{R}^{d \times 1} $ is the parameter. Assume that the mean squared error is used as the training loss, then solving {\eqref{optimal-SP}} is equivalent to solving
\begin{align}
\label{TPS-0}
      \min _{\bm \theta \in \mathbb{R}^{d} }   \norm{\bm y - \bm X \bm \theta}_2^2 + \gamma f(\bm \theta),
\end{align}
where $f(\bm \theta):= \| \bm \theta \|_1 +  \zeta \bar f( \bm \theta, {\bm w})$ with $\zeta = {\lambda}/{ \gamma }$. It is known that when $\zeta = 0$, \eqref{TPS-0} is reduced to the standard unconstrained compressed sensing problem under noisy conditions~\cite{donoho2006compressed,eldar2012compressed}, and the size of the tangent cone can be used to analyze the probability of obtaining the optimal sparse solution, \textcolor{black}{i.e., the smaller the tangent cone, the higher probability to get the optimal solution for the same amount of training data.}~\cite{zhang2018recovery,8006522}. 
We then analyze the change in the size of the tangent cone under different personalization constraints.
Note that the normal cone at the optimal solution $\bm \theta^{\star}$ is the polar of its tangent cone, which can be generated by using the sub-differential of the objective function at $\bm \theta^{\star}$. We hence hope to find a personalized constraint that allows the sub-differential to move to a position close to zero. At this time, the normal cone is the largest, while the tangent cone is the smallest.
This motivates us to maximize the correlation between the client model and the global model to obtain personalization capabilities, i.e., $f(\bm \theta) = f_1(\bm \theta) := \| \bm \theta \|_1 - {\zeta_1} \langle  \bm \theta , {\bm w} \rangle $, then we have the corresponding normal cone is 
\begin{align}
\mathcal{N}_{f_1} = \text{cone}\{ {\partial \| \bm \theta^{\star} \|_1 - \zeta_1  \bm w } \}.
\end{align}
{\color{black} Here, $\partial \| \bm \theta^{\star} \|_1$ denotes the sub-differential of $\|\cdot\|_1$, defined as
\begin{align}
    \partial \norm{\bm \theta^\star}_1=\{\bm z |z_i=\mbox{sign}(\theta_i^\star), \forall i \in I, \text{and}~|z_i|\le 1, \forall \ i \in I^c\},
\end{align}
where $I$ denotes the support of $\bm \theta^\star$, $I^c$ is its complement, and $\mbox{sign}(z)$ returns $1, 0, -1$ for $z>0$, $z=0$ and $z<0$, respectively.  Assume the global parameter $\bm w$ is close to the optimal personalized parameter $\bm \theta^\star$ and shares the same support, then the sub-differential $\partial {f_1}(\bm \theta^{\star})$ is
\begin{align}
    \partial {f_1}(\bm \theta^{\star})=\{\bm z |z_i=\mbox{sign}(\theta_i^\star)-\zeta_1 w_i, \forall \ i \in I, ~\text{and}~~|z_i|\le 1, \forall \ i \in I^c\}.
\end{align}

Compared with the case without personalized constraints (when $\zeta=0$ and ${\partial \| \bm \theta^{\star}\|_1 }$), the sub-differential $\partial {f_1}(\bm \theta^{\star})$ is much closer to zero. 
So it can increase the probability of obtaining the optimal sparse solution (see Figure~\ref{fig_superisorty} ({Left})).}


\begin{remark}
The sub-differential $\partial {f_1}(\bm \theta^{\star})$ is close to zero since it is known that the angle between a network $\bm w$ and its sign ${\rm sign}(\bm w)$ is usually very small after sufficient training~\cite{chao2020directional}, and the converged client model is not too far from the global model, we have ${\partial \| \bm \theta^{\star} \|_1 - \zeta  \bm w } \approx 0$ with a suitable $\zeta$.  The detailed theoretical analysis will be given in the next section. 
\end{remark}

In contrast, when the $\ell_2$-norm distance is used as the personalization constraint~\cite{mcmahan2017communication,t2020personalized,li2018federated}, i.e., $f(\bm \theta) = f_2(\bm \theta) := \| \bm \theta \|_1 + \frac{\zeta_2}{2} \|  \bm \theta - {\bm w} \|_2^2$, the corresponding normal cone is given by
\begin{align}
\mathcal{N}_{f_2} 
&= \text{cone}\{ {\partial \| \bm \theta^{\star} \|_1 + \zeta_2 (\bm \theta^{\star} - \bm w) } \}.
\end{align}
\textcolor{black}{and the corresponding sub-differential $\partial {f_2}(\bm \theta^{\star})$
\begin{align}
    \partial {f_2}(\bm \theta^{\star})=\{\bm z |z_i=\mbox{sign}(\theta_i^\star)+\zeta_2 (\theta_i^\star-w_i), \forall \ i \in I,  ~\text{and}~~|z_i|\le 1, \forall i \in I^c\}.
\end{align}}

Since the converged client model is not too far from the global model, i.e., $(\bm \theta^{\star} - \bm w)$ is small, the normal cone $\mathcal{N}_{f_2}$ does not move much (see Figure~\ref{fig_superisorty} ({Middle})) and hence the probability of obtaining the optimal sparse solution of \eqref{TPS-0} does not change much. That is, although the $\ell_2$-norm based personalization method can obtain personalization capabilities, it cannot help the client to train a better sparse model with a small amount of data.

\section{FedMac}\label{sec_FedMac}

\subsection{{\texttt{FedMac}}: Problem Formulation}

Instead of solving the problem in \eqref{conv-FL}, we aim to find a {\em sparse global model} $\bm w$ based on {\em sparse personalized models} $\bm \theta_{i},~i=1,...,N$ via maximizing correlation between $\bm w$ and $\bm \theta_{i}$ by minimizing
\begin{align}
\label{FedMac-1n}
    {{\texttt{FedMac:}}} \min _{{\bm w} \in \mathbb{R}^{d}}\left\{F(\bm w)\triangleq \frac{1}{N} \sum_{i=1}^{N} F_{i}(\bm w)\right\},  
    \end{align}
    where
    \begin{align}
    F_{i}(\bm w) &\triangleq \min _{\bm \theta_{i} \in \mathbb{R}^{d}}  \left\{  H_i(\bm\theta_i;\bm w)  + \frac{\lambda}{2} \| \bm w \|_2^2  +  \gamma_w \| \bm w\|_1 \right\}, \nonumber \\
    H_i(\bm\theta_i;\bm w) &\triangleq  \ell_{i} \left(\bm \theta_{i} \right)   +  \gamma \| \bm \theta_{i} \|_1 - \lambda\left \langle \bm \theta_{i}, {\bm w}\right \rangle .\nonumber
\end{align}
On the server side, since the client data only serve for the update of $\bm \theta_{i}$ and there is no data on the server to restrict $\bm w$, the global model ${\bm w}$ obtained by maximizing the correlation $\left \langle \bm \theta_{i}, \bm w \right \rangle$ may gradually become larger. We hence add a penalty term  $\frac{\lambda}{2} \| {\bm w} \|_2^2 $ to counteract the divergence of the global model.

\subsection{\texttt{FedMac}: Algorithm}

Next we propose the algorithm {\texttt{FedMac}}. To start with, we use a twice continuously differentiable approximation to replace the $\ell_{1}$-norm used in {\texttt{FedMac}}, i.e., replacing $\|\bm x\|_1$ by~\cite{7755794}
\begin{align}
 \phi_{\rho}(\boldsymbol{\bm x})
     &=\rho \sum_{n=1}^{d} \log \cosh \left(\frac{{x}_{n}}{\rho}\right), \notag
\end{align}
where $x_n$ is the $n$-th element in $\bm x$ and $\rho$ is a weight parameter, which controls the smoothing level.  And we have the $n$-th element of $\nabla\phi_{\rho}(\bm x)$ is $    \left[\nabla\phi_{\rho}(\bm x)\right]_n=\tanh \left(x_{n} / \rho\right)$.
Note that we use $\phi_{\rho}({\cdot})$ instead of $\|\cdot\|_1$ to exploit the sparsity in the proposed algorithm because it makes the loss function has continuous differentiable property, which enables us to analyze the convergence of the proposed algorithm and facilitates gradient calculation and back propagation in the network. 

Next, we present the following useful assumptions, which are widely used in FL gradient calculation and convergence analysis~\cite{karimireddy2020scaffold, fallah2020personalized, li2019communication, yu2019linear}. 

\begin{assumption}\label{assumption_cs} 
{\color{black}
(Strong convexity and smoothness)
Assume that $\ell_i(\cdot):\mathbb{R}^{d} \rightarrow \mathbb{R} $ is $(a)$ $\mu$-strongly convex or $(b)$ nonconvex and L-smooth on $\mathbb{R}^{d}$, where $\mathbb{R}^{d}$ is endowed with the $\ell_2$-norm. 
}
\end{assumption}

\begin{assumption}\label{assumption_bv} (Bounded variance)
The variance of stochastic gradients (sampling noise) in each client is bounded by
\begin{align}
    \mathbb{E}_{Z_i}\left[\left\|\nabla \tilde{\ell}_{i}\left(\bm w ; Z_{i}\right)-\nabla \ell_{i}(\bm w)\right\|^{2}_2\right] \leq \gamma_{\ell}^{2},\notag
\end{align}
where $Z_{i}$ is a training data randomly drawn from the distribution of client $i$.
\end{assumption}

On this basis, we propose the smoothness and strong convexity proportion of $F_{i}(\bm w)$ in Proposition~\ref{proposition_convex_analyse}, which is proved in the appendix.

\begin{proposition}
\label{proposition_convex_analyse}
{\color{black}
If $\ell_{i}$ is $\mu$-strongly convex or nonconvex with $L$-Lipschitz $\nabla \ell_i$ on $\mathbb{R}^{d}$ w.r.t. $\|\cdot\|_2$, then $F_i$ with 
\begin{align}
\nabla_{\bm w} F_i = \lambda (\bm w - \hat{\bm \theta}(\bm w)) + \gamma_w \nabla \phi_{\rho}(\bm w),
\end{align}
is $\mu_F$-strongly convex with $\mu_F=\gamma_w \mu_\phi + \lambda$  and $L_F$-smooth with $L_F=\lambda + \frac{\lambda^2}{\bar\mu} + \frac{\gamma_w}{\rho}$, for 
$$
\bar\mu := \left\{ \begin{array}{l}
\mu_\phi + \mu, ~\text{\rm{if Assumption~1 (a)~holds;}}\\
\mu_\phi - L, ~\text{\rm{if Assumption~1 (b)~holds, and~ $\mu_\phi \geq L$.}}
\end{array} \right.
$$
where $\mu_\phi$ is the strong convexity parameter of $\phi_\rho(\cdot)$.
}
\end{proposition}

Our algorithm updates $\bm \theta_i$ and $\bm w$ by alternately minimizing the two subproblems of {\texttt{FedMac}}. In particular, we first update $\bm \theta_i$ by solving 
\begin{align}
\label{A-theta}
    \bm{\hat\theta}_{i} = \arg\min _{\bm \theta_{i} \in \mathbb{R}^{d}}\left\{\ell_{i}\left(\bm \theta_{i}\right) 
    +\gamma\phi_{\rho}(\bm \theta_{i}) 
    -{\lambda}\left\langle \bm \theta_{i}, \bm w \right\rangle   \right\} ,
\end{align}
then update the corresponding $\bm w_i$ (named {\em local global model}) by solving
\begin{align}
\label{A-w}
    \bm{\hat w}_{i}= \min_{\bm w_i \in \mathbb{R}^{d}}\left\{   -   \lambda  \langle \bm{\hat \theta}_i , \bm w_i  \rangle     +  \frac{\lambda}{2} \| {\bm w_i} \|_2^2    +\gamma_w\phi_{\rho}(\bm w_{i}) \right\}.
\end{align}
After that, we upload $\bm{\hat w}_i,~i = 1,...,N$ to the server and aggregate them to get the updated $\bm w$.


\begin{algorithm}[tp]
\label{Algorithm_FedMac-S}
\begin{minipage}[!ht]{\linewidth}
	\caption{\small{\texttt{FedMac}}: Sparse Personalized Federated Learning via Maximizing Correlation Algorithm}
	\begin{tabular}{l}
{\bf{Server executes:}} \\
	\hspace{0.5cm}{Input} $T, R, S, \lambda, \eta, \gamma,\eta_p, \gamma_w, \beta, \rho, \bm w^{0}$ \\
	\hspace{0.5cm}\bf{for} $t=0,1,...,T-1$ \bf{do} \\
      	\hspace{1.0cm}\bf{for} $i=1,2,...,N$ \bf{in parallel do} \\
			\hspace{1.5cm}$\bm w_i^{t+1} \leftarrow \text{ClientUpdate}(i,\bm w^t)$  \\
		\hspace{1.0cm}$\mathcal{S}^{t} \leftarrow \text{(random set of $S$ clients)}$ \\
		\hspace{1.0cm}$\bm w^{t+1}=(1-\beta) \bm w^{t}+\frac{\beta}{S} \sum_{i \in S^t} {\bm w_i^{t+1}}$ \\
	\\
	$\textbf{ClientUpdate}(i,\bm w^{t})\textbf{:}$ \\
		\hspace{0.5cm}$\bm \theta_{i}^{t,0} = \bm w_{i}^{t,0} = \bm w^{t}$  \\
		    \hspace{0.5cm}{\bf{for}} $r=0,1,...,R-1$ \bf{do}  \\
		        \hspace{1.0cm}$\mathcal{D}_{i} \leftarrow \text{(sample a mini-batch with size $|\mathcal{D}|$ )}$ \\
		        \hspace{1.0cm}Update $\bm{\tilde\theta}_{i}^{t,r}$ according to \eqref{A-H-min}\\
		         \hspace{1.0cm}$\bm w_{i}^{t, r+1} = \bm w_{i}^{t,r}-\eta \nabla_{\bm w} F_{i}\left(\bm w_{i}^{t,r}\right)$\\
		\hspace{0.5cm}Return $\bm{\bm w}_{i}^{t,R}$ to the server
	\end{tabular}
\end{minipage}
\end{algorithm}

Note that \eqref{A-theta} can be easily solved by many first order approaches, for example the stochastic gradient descent~\cite{zinkevich2010parallelized}, based on the gradient
\begin{align}
   \nabla_{\bm \theta} F_{i} = \nabla \ell_{i}(\bm \theta_{i}) +\gamma \nabla \phi_\rho(\bm \theta_{i}) - \lambda  \bm w  \notag
\end{align}
with a learning rate $\eta_p$.
However, calculate the exactly $\nabla \ell_{i}(\bm \theta_{i})$ requires the distribution of $Z_{i}$, we hence use $\nabla \tilde \ell_{i}(\bm\theta_{i},\mathcal{D}_{i}) = \frac{1}{|\mathcal{D}_{i}|}\sum_{Z_{i} \in \mathcal{D}_{i}} \nabla  \ell_{i}(\bm\theta_{i},Z_{i})$ instead, i.e., we sample a mini-batch of data $\mathcal{D}_{i}$ to obtain the unbiased estimate of $\nabla \ell_{i}(\bm \theta_{i})$,
such that $\mathbb{E}[ \nabla \tilde \ell_{i}(\bm\theta_{i},\mathcal{D}_{i}) ] = \nabla \ell_{i}(\bm \theta_{i})$. Therefore, we solve the minimization problem 
\begin{align}
\label{A-H-min}
\bm{\tilde \theta}_{i}^{t,r}(\bm{ w}_{i}^{t,r}) = \arg\min _{\boldsymbol{\theta}_{i} \in \mathbb{R}^{d}} \tilde H(\bm \theta_{i};\bm{ w}_{i}^{t,r},\mathcal{D}_{i}),
\end{align}
instead of solving \eqref{A-theta} to obtain an approximated personalized client model, where  $\bm{ w}_{i}^{t,r}$ is the current local global model w.r.t. the $i$-th client, $t$-th global round, and $r$-th client round, $\bm{\tilde \theta}_{i}^{t,r}$ is the corresponding estimated client model and 
\begin{align}
\tilde H(\bm \theta_{i};\bm{w}_{i}^{t,r},\mathcal{D}_{i}) = \tilde \ell_{i}\left(\boldsymbol{\theta}_{i},\mathcal{D}_{i} \right)  +\gamma \phi_{\rho}(\bm \theta_{i}) -  \lambda \left\langle \bm \theta_{i}, \bm{w}_{i}^{t,r} \right\rangle . 
\end{align}
Similarly, \eqref{A-H-min} can be solved by the stochastic gradient descent. We let the iteration go until the condition $\| \nabla \tilde H(\bm \theta_{i};\bm{w}_{i}^{t,r},\mathcal{D}_{i}) \|^2_2 \leq \nu$ is reached, where $\nu$ is an accuracy level.

Once the client model is updated, the corresponding local global model is updated by stochastic gradient descent as follows
\begin{align}
    \bm w_{i}^{t, r+1} = \bm w_{i}^{t,r}-\eta \nabla_{\bm w} F_{i}\left(\bm w_{i}^{t,r}\right),
\end{align}
where $\eta$ is a learning rate. Finally, we summarize our algorithm in Algorithm~1. Similar to~\cite{t2020personalized,karimireddy2020scaffold},  an additional parameter $\beta$ is used for global model update to improve convergence performance, and we average the global model over a subset of clients $\mathcal{S}^t$ with size $S$ to reduce the occupation of bandwidth.

\section{Theoretical Analysis}\label{sec_theoretical_analysis}

\subsection{Convergence Analysis}

For unique solution $\bm w^{\star}$ to {\texttt{FedMac}}, which always exists for strongly convex $F_i$, we have the following important Lemmas and Theorem~\ref{theorem_convergence_FedMac}, which are proved in Appendix.

\begin{lemma}[Bounded diversity of $\bm \theta_i$ w.r.t. mini-batch sampling]
\label{lemma_theta-theta}
Let  $\tilde{\bm \theta}_i (\bm w_i^{t,r})$ be a solution to $\| \nabla \tilde{H}_i ( \tilde{\bm \theta}_i; \bm w_i^{t,r}, \mathcal{D}_i ) \|_2^2 \leq \nu $, if Assumptions~\ref{assumption_cs} and \ref{assumption_bv} hold, we have
$$ \mathbb{E}\left[ \| \tilde{\bm \theta}_i(\bm w_i^{t,r}) - \hat{\bm \theta}_i(\bm w_i^{t,r}) \|_2^2  \right] \leq \delta^2 = \frac{2}{\mu^2} \left( \frac{\gamma^2_\ell}{|D|} + \nu \right).$$
\end{lemma}

\begin{lemma}[Bounded client drift error]
\label{lemma_g-F}
 If $\eta \leq \frac{1}{2L_F\sqrt{R(1+R)}} \Leftrightarrow \tilde\eta \leq \frac{\beta \sqrt{R}}{2L_F\sqrt{1+R}}  $ and Assumptions~\ref{assumption_cs} and \ref{assumption_bv} hold, we have
\begin{align}
    \frac{1}{N R} \sum_{i, r=1}^{N, R} \mathbb{E}\left[\left\|g_{i}^{t,r} - \nabla F_{i}\left(\bm w^{t}\right)\right\|_2^{2}\right] \notag
    \leq&~   64\tilde\eta L_F^2  \mathbb{E} \left[     F\left(\bm w^{t}\right)  -  F\left(\bm w^{\star}\right)     \right] +  8\sigma_{F}^{2}  +  10 \lambda^{2} \delta^{2}, \notag
\end{align}
where $g_{i}^{t,r} = \lambda\left(\bm w_{i}^{t,r}-\tilde{\bm \theta}_{i}(\bm w_{i}^{t,r})\right)+\gamma_w\nabla\phi_\rho(\bm w_{i}^{t,r})$ and $\sigma_{F}^{2} \triangleq   \frac{1}{N} \sum_{i}^{N}\left\|\nabla F_{i}\left(\bm w^{\star}  \right)  \right\|_2^{2}$.
\end{lemma}

\begin{lemma}[Bounded diversity of $F_i$ w.r.t. client sampling]  If Assumption~\ref{assumption_cs} holds, we have
\label{lemma_Fi-F}
\begin{align}
	\mathbb{E}_{S_t}\left\|\frac{1}{S} \sum_{i \in \mathcal{S}^{t}} \nabla F_{i}\left(\bm w^{t}\right)-\nabla F\left(\bm w^{t}\right)\right\|_2^{2} 
	\leq&~ \frac{N/S-1}{N(N-1)}\sum_{i=1}^{N}\left( \left\|\nabla F_{i}\left(\bm w^{t}\right)-\nabla F\left(\bm w^{t}\right)\right\|_2^{2}\right). \notag
\end{align}
\end{lemma}

\begin{lemma}[Bounded diversity of $F_i$ w.r.t. distributed training]  If Assumption~\ref{assumption_cs} holds, we have
\label{lemma_Fi-F_s}
$$  \frac{1}{N} \!\sum_{i=1}^{N}\left\|\nabla F_{i}(\bm w) \!- \nabla F(\bm w)\right\|_2^{2} \leq 4 L_{F}\left(F(\bm w) \!- F\left(\bm w^{\star}\right)\right) \!+ 2\sigma_{F}^{2}, $$
where $\sigma_{F}^{2} \triangleq   \frac{1}{N} \sum_{i}^{N}\left\|\nabla F_{i}\left(\bm w^{\star}\right)\right\|_2^{2}$.
\end{lemma}

\begin{lemma}[One-step global update]
\label{lemma_one_step}
If Assumption~\ref{assumption_cs} holds, we have
\begin{align}
    \mathbb{E}\left[\left\|\bm w^{t+1}-\bm w^{\star}\right\|_2^{2}\right] \notag
    \leq&~ \mathbb{E}\left[\left\|\bm w^{t}-\bm w^{\star}\right\|_2^{2}\right]-\tilde{\eta}\left(2-6 L_{F} \tilde{\eta}\right)\mathbb{E}\left[F\left(\bm w^{t}\right)-F\left(\bm w^{\star}\right)\right]\notag\\
    &~ +\frac{\tilde{\eta}\left(3 \tilde{\eta}+1 / \mu_{F}\right)}{N R} \sum_{i, r}^{N, R} \mathbb{E}\left[\left\|g_{i}^{t,r}-\nabla F_{i}\left(\bm w^{t}\right)\right\|^{2}_2\right]\notag\\
    &~ + 3 \tilde{\eta}^{2}\mathbb{E}\left[\left\|\frac{1}{S} \sum_{i \in \mathcal{S}^{t}}\nabla F_{i}\left(\bm w^{t}\right)-\nabla F\left(\bm w^{t}\right)\right\|_2^{2}\right].  \notag
\end{align}
\end{lemma}

\begin{remark}
Lemma~\ref{lemma_theta-theta} shows the diversity of $\bm \theta_i$  w.r.t. mini-bath sampling is bounded. Lemma~\ref{lemma_g-F} shows the client drift error caused by mini-batch training and local update is bounded. Lemma~\ref{lemma_Fi-F} and \ref{lemma_Fi-F_s} show the diversities of $F_i$ w.r.t. client sampling and distributed training are bounded. Using all the above lemmas, we can get the error bound 
of the one-step update of the global model in Lemma~\ref{lemma_one_step}.
\end{remark}

\begin{theorem}
\label{theorem_convergence_FedMac}
Let Assumptions~\ref{assumption_cs} and \ref{assumption_bv} hold. If $\eta \leq \frac{\hat\eta}{\beta R}$, where $\hat\eta = \frac{1}{(18+256\kappa_F)L_F}$ and $\beta \geq 1$, then we have 
\begin{align}
    (a)~&~\frac{1}{T} \sum_{t=0}^{T-1}  \mathbb{E} \left [F(\bm w^t) -F(\bm w^{\star})  \right] \leq \mathcal{O}(\Delta)  \notag 
     \triangleq ~ \mathcal{O}    \left\{   \frac{\Delta_0}{\hat\eta T} + \frac{ \sigma_F^2 }{\mu_F} + \frac{ \lambda^2\delta^2}{\mu_F} + \frac{ (\sigma_F^2 \Delta_0(N/S-1))^{1/2}}{\sqrt{TN}}    \right\},    \notag\\
    (b)~&~\frac{1}{N} \sum_{i=1}^N \mathbb{E} \left[  \|  \tilde{\bm \theta}^T_i(\bm w^T)   -  \bm w^{\star} \|^2_2  \right]   \notag 
     \leq ~ \frac{L_F^2+\lambda^2}{\lambda^2\mu_F} \mathcal{O}    (  \Delta )    +  \mathcal{O} \left(  \delta^2 + \frac{\sigma_F^2 + \gamma_w^2 d_s^2}{\lambda^2} \right) , \notag
\end{align}
where $\sigma_{F}^{2} \triangleq   \frac{1}{N} \sum_{i=1}^{N}\left\|\nabla F_{i}\left(\bm w^{\star}\right)\right\|_2^{2}$,  $\delta^2 = \frac{2}{\mu^2} \left( \frac{\gamma^2_\ell}{|D|} + \nu \right)$, $\Delta_0 \triangleq \mathbb{E} \left[  \| \bm w^0 - \bm w^{\star}  \|^2_2 \right]$ and $d_s$ denotes the number of non-zero elements in $\bm w$.
\end{theorem}

\begin{remark}
Theorem~\ref{theorem_convergence_FedMac}(a) shows the convergence of the global model. The first term is caused by the initial error $\Delta_0$, which decreases linearly with the increase of training iterations. The second term is caused by client drift with multiple local updates. The third term shows that {\texttt{FedMac}} converges towards a $\frac{ \lambda^2\delta^2}{\mu_F}$-neighbourhood of $\bm w^{\star}$. The last term is due to the client sampling, which is 0 when $S=N$. We can see that the sparse constraints in {\texttt{FedMac}} do not affect the convergence rate of the global model. Theorem~\ref{theorem_convergence_FedMac}(b) shows the convergence of personalized models in average to a ball of center $\bm w^{\star}$ and radius $\mathcal{O} \left\{  \delta^2 + \frac{\sigma_F^2 + \gamma_w^2 d_s^2}{\lambda^2} + \frac{ \lambda^2\delta^2}{\mu_F}\right\} $, which means that a less sparse model requires a larger $\lambda$ to strengthen the connection between the client models and the global model. \textcolor{black}{Note that the limitation on the learning rate doesn't increase the training time according to our experiments, which is useful and common for Federated Learning analysis. Similar to works in~\cite{t2020personalized, karimireddy2020scaffold}, we apply this technique to our theorem analysis. }
\end{remark}

\subsection{Theoretical Performance Superiority}

In this subsection, we first present some useful definitions, then present the performance superiority of {\texttt{FedMac}} based on Assumptions~\ref{assumption_cs}-\ref{assumption_l1}. Suppose that the mean squared error is used as the training loss, i.e., $\ell(\bm \theta) = \norm{\bm y - \bm X \bm \theta}_2^2$. 
For fixed $\bm w$, we have the following Theorem~\ref{theorem_Base}, which is proved in Appendix.

\begin{definition}
A random variable $\theta$ is called \emph{sub-Gaussian} if it has finite Orlicz norm
$$
\norm{\theta}_{\psi_2}=\inf \{t>0: \E \exp(\theta^2/t^2)\le 2\},
$$
where $\norm{\theta}_{\psi_2}$ denotes the sub-Gaussian norm of $\theta$.  In particular,  Gaussian, Bernoulli and all bounded random variables are sub-Gaussian. 
\end{definition}

\begin{definition}
A random vector $\bm \theta \in \R^{N_I}$ is called \emph{sub-Gaussian} if $\left\langle \bm \theta,\vw \right\rangle$ is sub-Gaussian for any $\vw \in \R^{N_I}$, and its sub-Gaussian norm is defined as
$$
\norm{\bm \theta}_{\psi_2} = \sup \limits_{\vw \in \S^{N_I-1}} \norm{\left\langle \bm \theta,\vw \right\rangle}_{\psi_2}.
$$
\end{definition}

\begin{definition}
A random vector $\bm \theta \in \R^{N_I}$ is \emph{isotropic} if it satisfies $\E \bm \theta \bm \theta^T = \bm{I}_{N_I}$, where $\bm{I}_{N_I} \in \R^{N_I \times N_I}$  is the identity matrix.
\end{definition}

\begin{definition}
The \emph{subdifferential} of a convex function $f: \R^{N_I} \to \R$ at $\bm \theta^\star$ for all $\vd \in \R^{N_I}$ is defined as the set of vectors
$$
	\partial f(\bm \theta^\star) = \{\vu \in \R^{N_I}: f(\bm \theta^\star + \vd) \geq f(\bm \theta^\star) + \langle \vu, \vd \rangle.
	$$
	\end{definition}

\begin{definition}
The \emph{Gaussian width} of a subset $\mathcal{E} \subset \R^{N_I}$ is defined as
$$
w(\mathcal{E})= \E \sup \limits_{\bm \theta \in \mathcal{E}} \ip{\vg}{\bm \theta},~\vg \sim N(0,\bm{I}_{N_I}),
$$
and  the \emph{Gaussian complexity} of a subset $\mathcal{E} \subset \R^{N_I}$ is defined as
$$
\xi (\mathcal{E})= \E \sup \limits_{\bm \theta \in \mathcal{E}} |\ip{\vg}{\bm \theta}|,~\vg \sim N(0,\bm{I}_{N_I}).
$$
These two geometric quantities have the following relationship \cite{Chen2019}
\begin{equation}\label{Relation}
	\xi(\mathcal{E}) \le  2 w(\mathcal{E})+\norm{\vy}_2~~~\textrm{for every}~\vy \in \mathcal{E}.
\end{equation}
\end{definition}

\begin{definition}
The \emph{Gaussian squared distance} $\eta(\mathcal{E})$ is defined as
$$
\eta^2 (\mathcal{E})= \E \inf \limits_{\bm \theta \in \mathcal{E}} \norm{\vg-\bm \theta}_2^2,~\vg \sim N(0,\bm{I}_{N_I}).
$$
\end{definition}

\begin{definition}
Define the error set 
\begin{equation} \label{def: error_set}
	\mathcal{E}_f=\{\vd \in \R^{N_I}: f(\bm \theta^\star+ \vd) \le f(\bm \theta^\star)\},
\end{equation}
then it belongs to the following convex set
\begin{equation} \label{def: convex_cone}
	\mathcal{C}_f =\{\vd \in \R^{N_I}: \ip{\vd}{\vu} \le  0~~\textrm{for any}~\vu \in \partial f(\bm \theta^\star)\}.
\end{equation}
\end{definition}

\begin{assumption}\label{assumption_net}
Assuming that the network is a single-layer network, such that $\bm y = \bm X \bm \theta^\star$ with solution $\bm \theta^{\star} \in \mathbb{R}^{d \times 1} $.
\end{assumption}

\begin{assumption}\label{assumption_data}
Assuming that $\vX \in \R^{N_D \times N_I}$ is a random matrix whose rows $\{\vX_{i}\}_{i=1}^{N_D}$ are independent, centered, isotropic and sub-Gaussian random vectors.
\end{assumption}

\begin{assumption}\label{assumption_sign}
After sufficient training iterations $T$, there exists a constant $\zeta>0$ such that $\zeta\bm w - {\rm sign}(\bm w) \approx 0$ and $\zeta\bm \theta - {\rm sign}(\bm \theta) \approx 0$.
\end{assumption}

\begin{assumption}\label{assumption_l1}
Define $I=\{n:  \theta_n^\star \neq 0\}$ and $J=\{n:  \theta_n^\star \neq w_n\}$, assume that $N_I \gg q$ with $q=|I \cup J|$.
\end{assumption}

\begin{remark} 
{\rm{\textbf{[Condition (A3)]:}}} 
{\color{black}Neural networks are black boxes, and it is difficult to do quantitative analysis due to the high nonlinearity. Although Assumption~3 is a strong assumption, we only use it for quantitative analysis of superiority. We verify that maximizing correlation is better than $\ell_2$-norm based methods to guide sparse solutions under linear model constraints, which is heuristic and induces the proposition of our algorithm {\texttt{FedMac}}. Finally, the effectiveness of the {\texttt{FedMac}} for general nonlinear models are verified by simulation experiments.}
{\rm{\textbf{[Condition (A4)]:}}} 
{\color{black}Assumption~\ref{assumption_data} is standard for compressed sensing analysis and is easily satisfied due to the random initialization of the neural network.}
{\rm{\textbf{[Condition (A5)]:}}} 
Empirically, Assumption~\ref{assumption_sign} holds as the Figure~2 in~\cite{chao2020directional} shows that the angle between a network parameter $\vw$ and its sign ${\rm{sign}}(\vw)$ is usually very small after sufficient training.
{\rm{\textbf{[Condition (A6)]:}}} 
Assumption~\ref{assumption_l1}  can be understood as we assume that the network is sparse, for which it is not difficult to obtain $N_I \gg q$.
\end{remark}

\begin{theorem}\label{theorem_Base}  Let Assumptions~\ref{assumption_net} and \ref{assumption_data} hold. Let $\hat{\bm \theta}$ be the solution of the following problem
\begin{align}
\label{TPS-1T}
      \min _{\bm \theta \in \mathbb{R}^{d} }   \norm{\bm y - \bm X \bm \theta}_2^2 + \gamma f(\bm \theta),
\end{align}
and $f(\cdot)$ satisfies  $|f(\bm \theta^\star)-f(\hat{\bm \theta})|\le \alpha_f ||\bm \theta^\star-\hat{\bm \theta}||_2$ for some $\alpha_f>0$.  Let $\mathcal{C}_f$ denote the convex set $\mathcal{C}_f =\{\vd \in \R^{N_I}: \ip{\vd}{\vu} \le  0~~\textrm{for any}~\vu \in \partial f(\bm \theta^\star)\}$. If  $\gamma \leq 1 / \alpha_f$ and the number of training data satisfies
	\begin{equation}\label{NumberofMeasurements}
		\sqrt{N_D} \ge CK^2 \xi(\mathcal{C}_f \cap \S^{N_I-1})+ \epsilon,
	\end{equation}
	then with probability at least $1- 2\exp(-\xi^2(\mathcal{C}_f \cap \S^{N_I-1}))$, the solution $\hat{\bm \theta}$ satisfies
	\begin{equation}
	\| \bm \theta^\star-\hat{\bm \theta}\|_2 \le \frac{1}{\epsilon^2},
    \end{equation}
	where $\epsilon,C$ are absolute constants, $\xi (\mathcal{E})= \E \sup_{\bm \theta \in \mathcal{E}} |\ip{\vg}{\bm \theta}|$ denotes the Gaussian complexity with $\vg \sim N(0,\bm{I}_{N_I})$ and $K=\max_i \norm{\vX_{i}}_{\psi_2}$ denotes the maximum sub-Gaussian norm of $\{\vX_{i}\}_{i=1}^{N_D}$.	
\end{theorem}

\begin{remark}
Theorem~\ref{theorem_Base} indicates that once Assumptions~\ref{assumption_net} and \ref{assumption_data} hold, by using a small $\gamma$ such that $\gamma \leq 1 / \alpha_f$ and enough training data satisfying \eqref{NumberofMeasurements}, we can obtain the robust estimation of the sparse parameters of the network with high probability by minimizing the loss function \eqref{TPS-1T}.
\end{remark}

Next, we apply Theorem \ref{theorem_Base} to two sparse cases $f_1(\bm \theta) =   \| \bm \theta \|_1 -  \zeta_1 \left \langle \bm \theta, {\bm w}\right \rangle $ and $f_2(\bm \theta) =  \| \bm \theta \|_1  +  \zeta_2/2 \norm{\bm \theta - {\bm w}}_2^2 $ to compare their performance by calculating the upper bounds of $\xi(\mathcal{C}_{f_1} \cap \S^{N_I-1})$ and $\xi(\mathcal{C}_{f_2} \cap \S^{N_I-1})$, respectively. The results are presented in Theorem~\ref{UpperBound}, which is proved in the appendix.


\begin{theorem} \label{UpperBound} Let Assumptions~\ref{assumption_cs}-\ref{assumption_l1} hold, then we have
\begin{align}
  \xi(\mathcal{C}_{f_1} \cap \S^{N_I-1}) &\le  \mathcal{O} ( \sqrt{v_{\zeta_1} \log ({N_I})	}   ), \notag \\
  \xi(\mathcal{C}_{f_2} \cap \S^{N_I-1}) &\le  \mathcal{O} ( \sqrt{v_{\zeta_2} \log ({N_I})	}   ), \notag
\end{align}
where
$v_{{\zeta_1} } = \sum_{i \in I}({\rm{sign}}(\bm \theta_i^\star)-{\zeta_1} \vw_i)^2+\sum_{i \in K_{\zeta_1}^{\neq}} ({\zeta_1} |\vw_i|-1)^2$,
$v_{\zeta_2}= \sum_{i \in I}({\rm{sign}}(\bm \theta_i^\star)+\zeta_2 (\bm \theta_i^\star-\vw_i))^2+\sum_{i \in K_{\zeta_2}^{\neq}} (\zeta_2 |\vw_i|-1)^2,
$
$I=\{i: \bm \theta_i^\star \neq 0\}$, $J=\{i: \bm \theta_i^\star \neq \vw_i\}, $
and $ K_\zeta^{\neq}=\{ i \in I^c\cap J: |\vw_i|>1/{\zeta}\}$. 

Moreover, if $\eta \leq \frac{\hat\eta}{\beta R}$, $\bm w = \bm w^{\star}$ and $\gamma \leq 1 / \alpha_f$ we further have
\begin{align}
v_{{\zeta_1} } \leq \Delta^v(\zeta_1), \quad
v_{\zeta_2} \leq 2 |I| + \Delta^v(\zeta_2) , \notag
\end{align}
where $\Delta^v(\zeta) = {2 {\zeta}^2} \left( {2\gamma_w^2 d_s^2}/{\lambda^2} +  {1}/{\epsilon^4}  \right) $.
\end{theorem}

\begin{remark}
Theorem~\ref{UpperBound} indicates that the difference between the upper bounds of $\xi(\mathcal{C}_{f_1} \cap \S^{N_I-1})$ and $\xi(\mathcal{C}_{f_2} \cap \S^{N_I-1})$ is determined by the difference between $v_{{\zeta_1} }$ and $v_{{\zeta_2} }$. Note that from the components in $\Delta^v(\zeta)$, it can be seen that it mainly represents the error after the algorithm converges, so it is a relatively small value, especially when $\zeta$ is small. Therefore, the training data required for \eqref{TPS-1T} using $f_1(\bm \theta)$ is much less than that required for \eqref{TPS-1T} using $f_2(\bm \theta)$.
\end{remark}


\section{Experimental Results}\label{sec_experiment_results}

\subsection{Experimental Setting}\label{sec_set}

The proposed $\texttt{FedMac}$ is a personalized FL based on local fine-tuning, so we compare the performance of {\texttt{FedMac}} with FedAvg~\cite{mcmahan2017communication} and local fine-tuning personalized FL methods, including
Fedprox~\cite{li2018federated},
Per-FedAvg~\cite{NEURIPS2020_24389bfe},
HeurFedAMP~\cite{huang2021personalized} and pFedMe~\cite{t2020personalized}, on non-i.i.d. datasets. In communication cost simulations, we only compare \texttt{FedMac} with methods based on sparse constraints. Since other methods (e.g. ternary compression~\cite{8889996}, asynchronous learning~\cite{8945292} and quantization~\cite{9305988}) that can reduce the amount of communication can be combined with methods based on sparse constraints

We generate the non-i.i.d. datasets based on four public benchmark datasets, MNIST~\cite{lecun2010mnist,lecun1998gradient}, FMNIST (Fashion-MNIST)~\cite{xiao2017fashion}, CIFAR-100~\cite{krizhevsky2009learning} and Synthetic datasets \cite{li2018federated}. For MNIST, FMNIST and CIFAR-100 datasets, we follow the non-i.i.d. setting strategy in ~\cite{t2020personalized}. Each client occupies a unique local data with different data sizes and only has 2 of the 10 labels. 
The number of clients for MNIST/FMNIST is $N=20$ and the number of clients for CIFAR-100 is $N=10$. For the Synthetic dataset, we follow the non-i.i.d. setting strategy in \cite{li2018federated} for $N=100$ clients by setting $\bar \alpha=\bar\beta=0.5$ to control the differences in the local model and dataset of each client.

We set $S=10$ for experiments on MNIST and FMNIST datasets, while set $S=2$ and $S=20$ for experiments on CIFAR-100 and Synthetic datasets, respectively. A two-layer deep neural network with a hidden layer size of 100 is used for experiments on MNIST and FMNIST datasets, and a hidden layer of size 20 is used on the Synthetic dataset. A VGG-Net~\cite{simonyan2014very} is used for experiments on CIFAR-100 dataset. 
\textcolor{black}{For all algorithms, we follow the testing strategy in \cite{huang2021personalized} using $\ell_i(\bm \theta_i)$ to calculating training loss and evaluate the performance through the highest mean testing accuracy in all communication rounds of training.}

We did all experiments in this paper using servers with a GPU (NVIDIA Quadro RTX 6000 with 24GB memory), two CPUs (each with 12 cores, Inter Xeon Gold 6136), and 192 GB memory. The base DNN and VGG models and federated learning environment are implemented according to the settings in \cite{t2020personalized}. In particular, the DNN model uses one hidden layer, ReLU activation, and a softmax layer at the end. For the MNIST dataset, the size of the hidden layer is 100, while that is 20 for the Synthetic dataset. The VGG model is implemented for CIFAR-100 dataset with ``[16, `M', 32, `M', 64, `M', 128, `M', 128, `M']" cfg setting. We use PyTorch for all experiments.

For a fair comparison, we allow the stochastic gradient descent (SGD) algorithm used to solve \eqref{A-H-min} in Algorithm~1, i.e.,
$$
\bm{\tilde \theta}_{i}^{t,r}(\bm{ w}_{i}^{t,r}) = \arg\min _{\boldsymbol{\theta}_{i} \in \mathbb{R}^{d}} \tilde H(\bm \theta;\bm{ w}_{i}^{t,r},\mathcal{D}_{i}),
$$
to be iterated only once to make it consistent with the settings in FedAvg, Fedprox and Per-FedAvg. And we also modify HeurFedAMP and pFedMe to make the corresponding algorithms iterate only once.
The code based on PyTorch 1.8 will be available online.

\subsection{Effect of Hyperparameters}\label{sec_result_hyper}

\begin{figure}[!ht]
	\centering
	\subfloat{
	\includegraphics[width=2.5in]{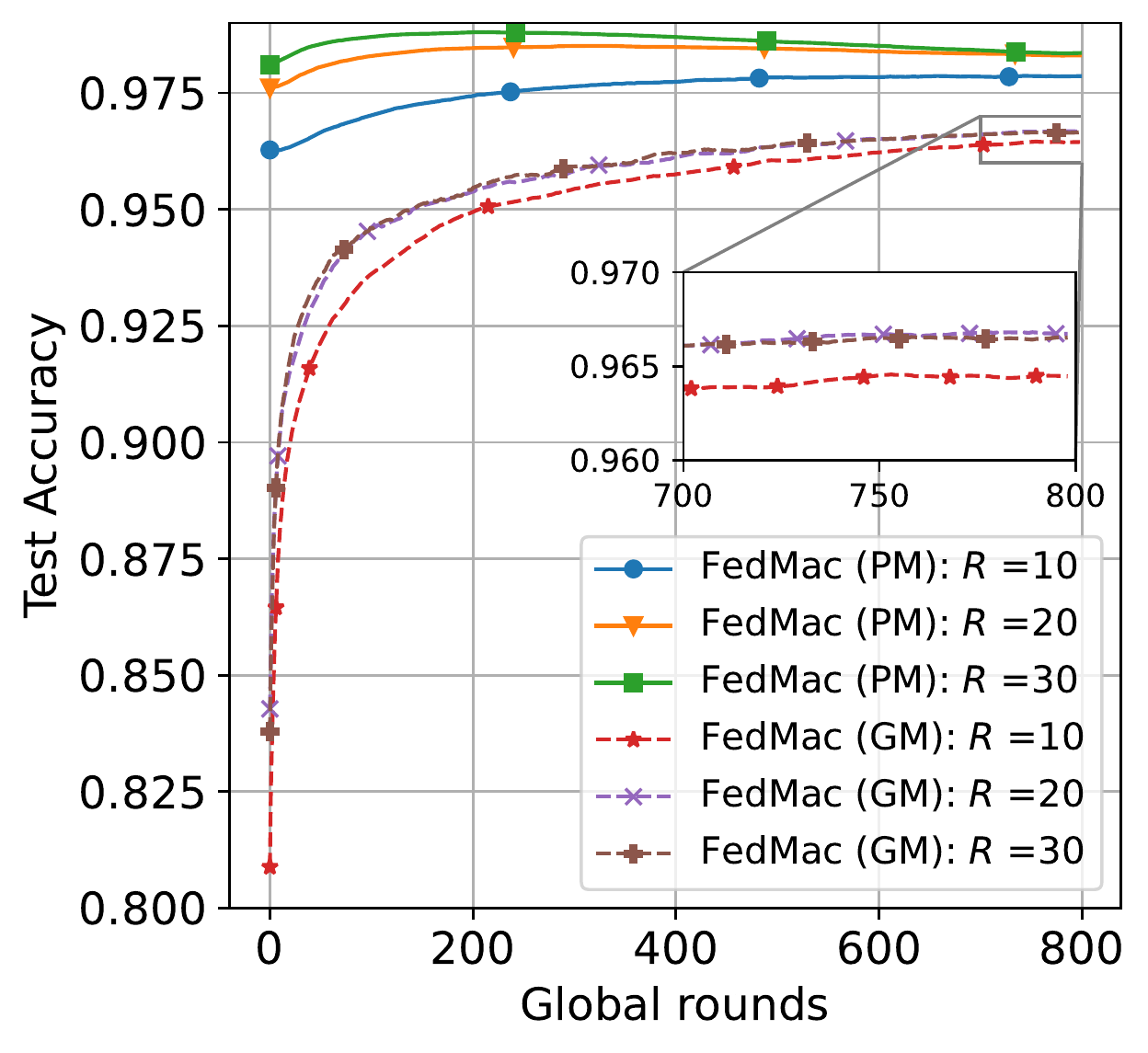}
	}
	\subfloat{
	\includegraphics[width=2.5in]{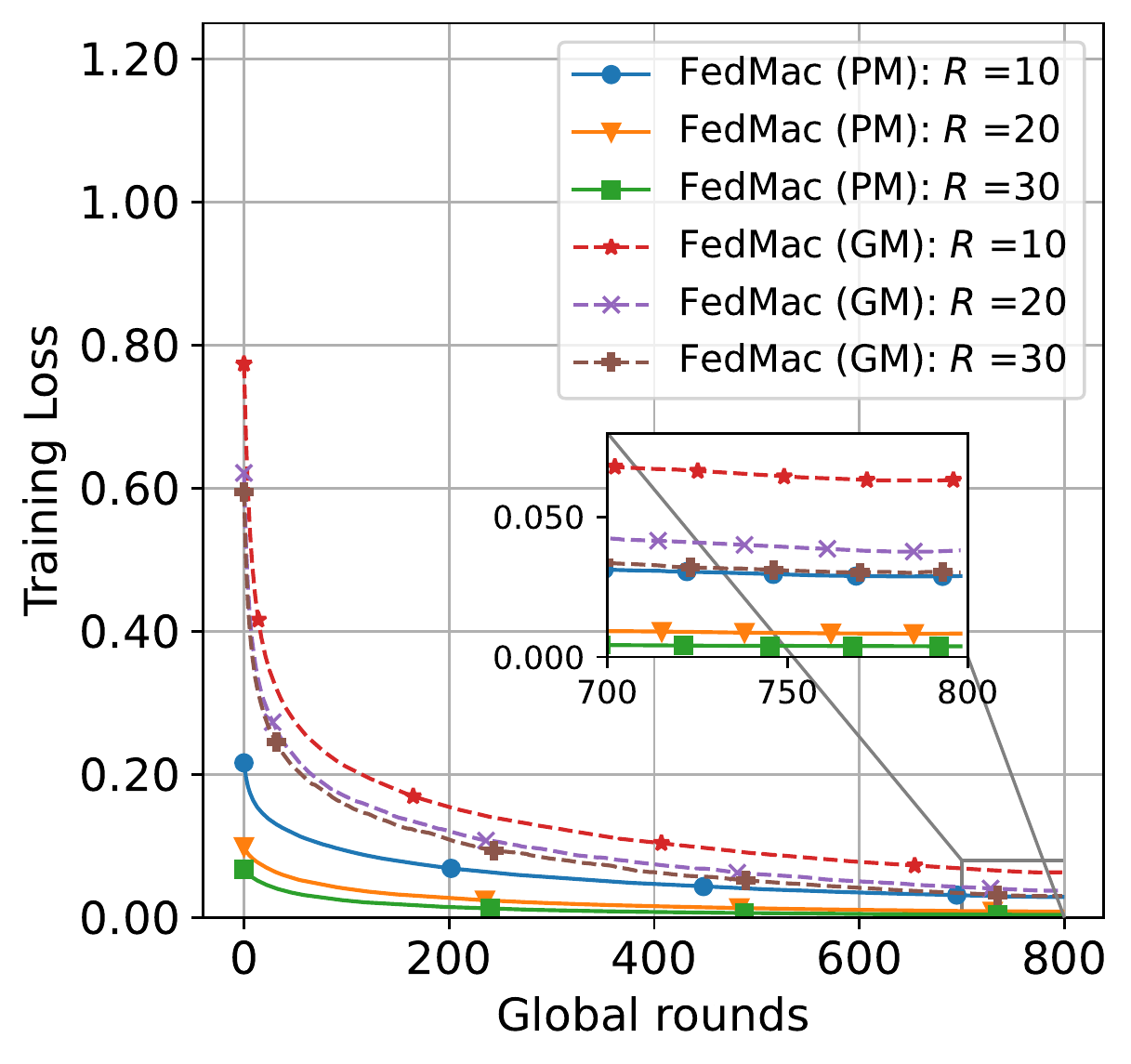}
	}
	
	\subfloat{
	\includegraphics[width=2.5in]{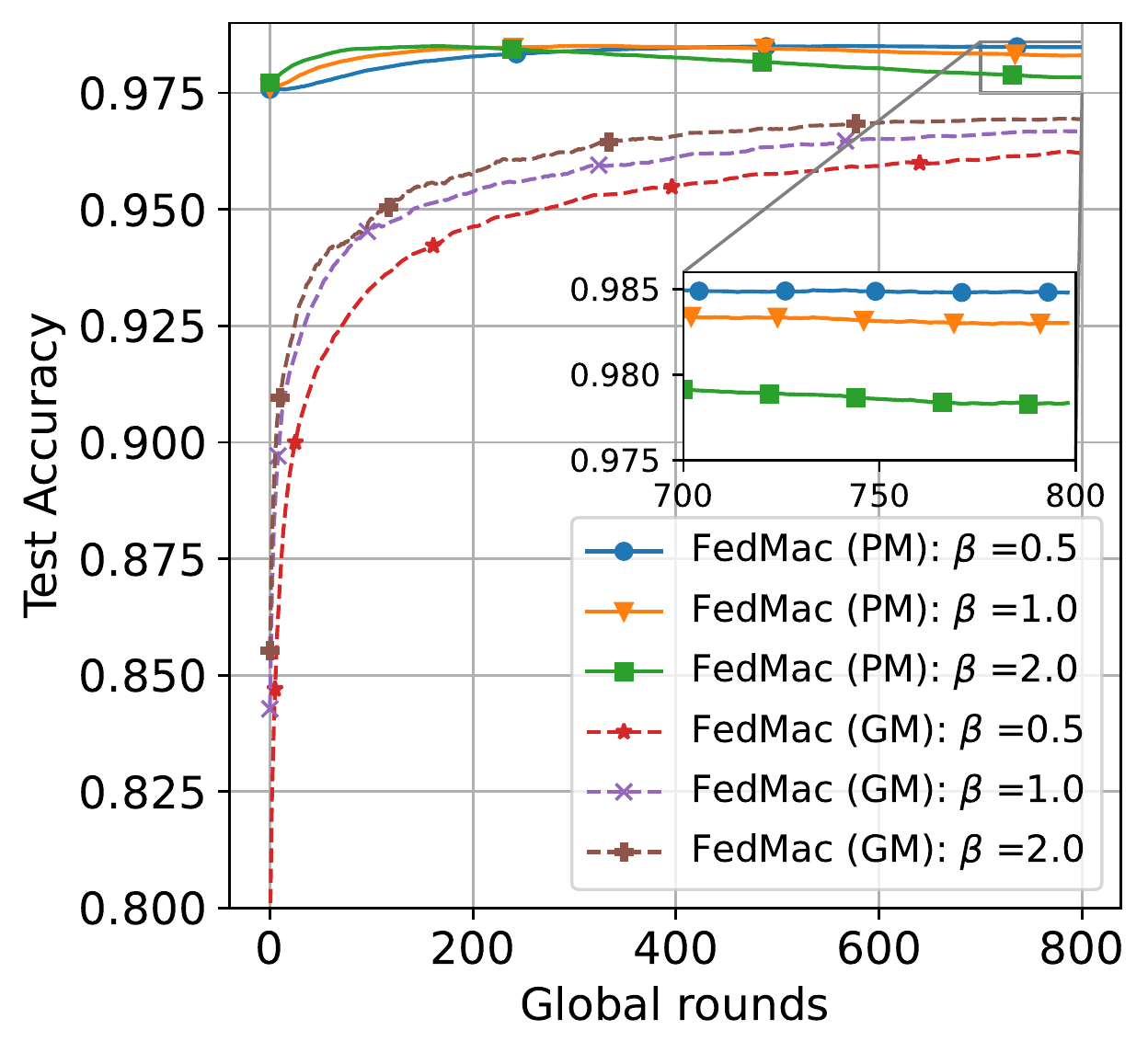}
	}
	\subfloat{
	\includegraphics[width=2.5in]{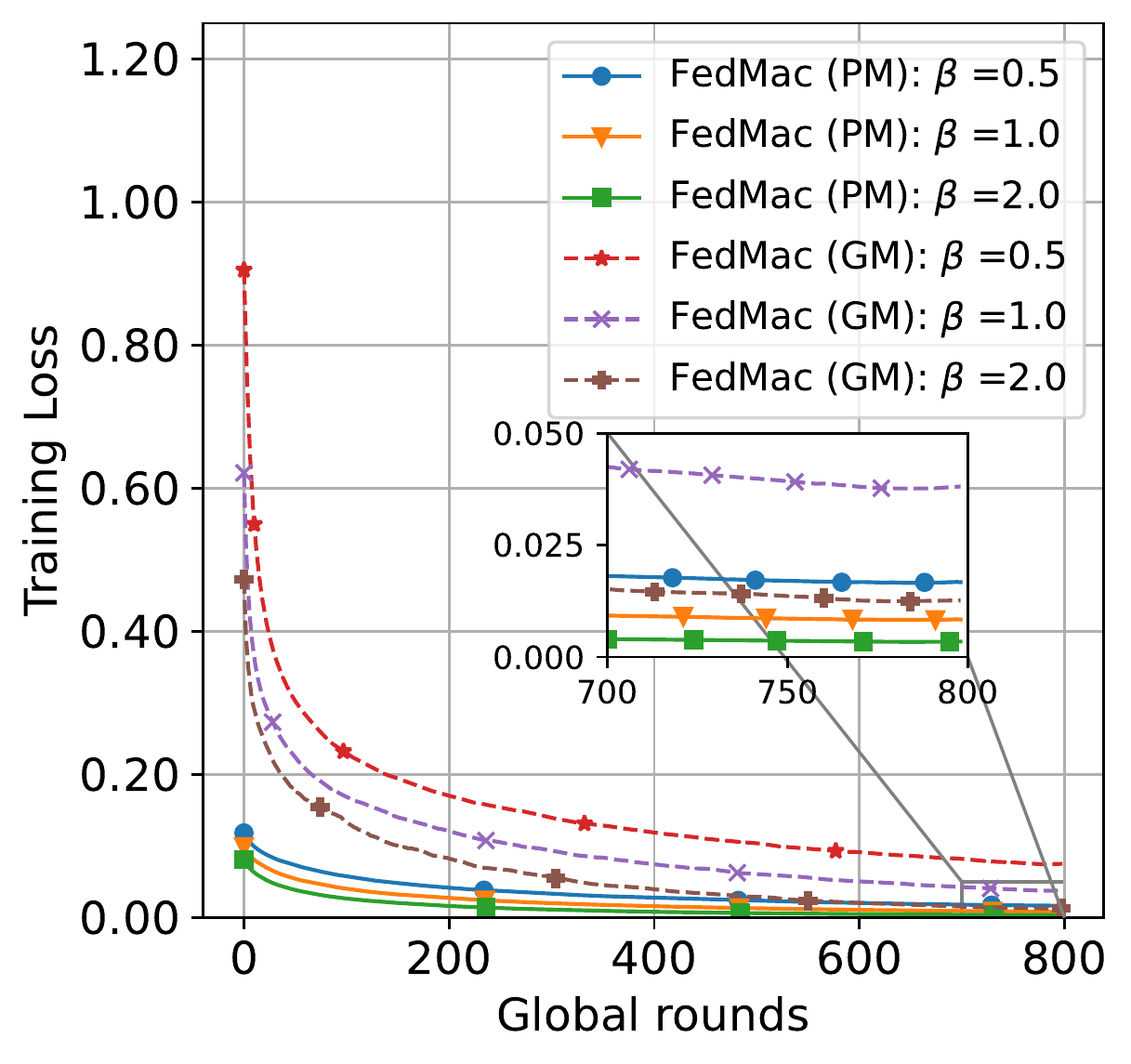}
	}

    \subfloat{
	\includegraphics[width=2.5in]{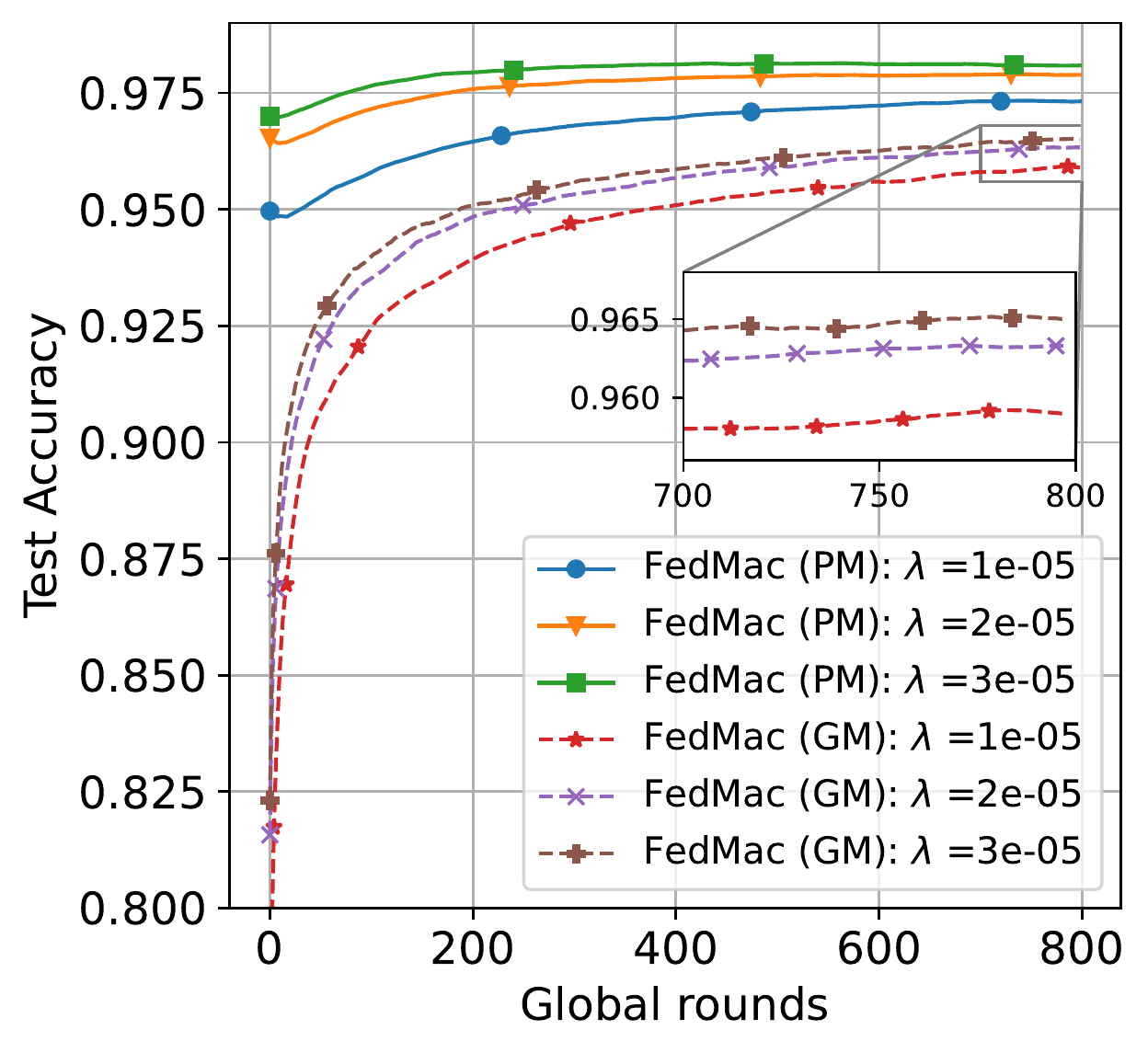}
	}
	\subfloat{
	\includegraphics[width=2.5in]{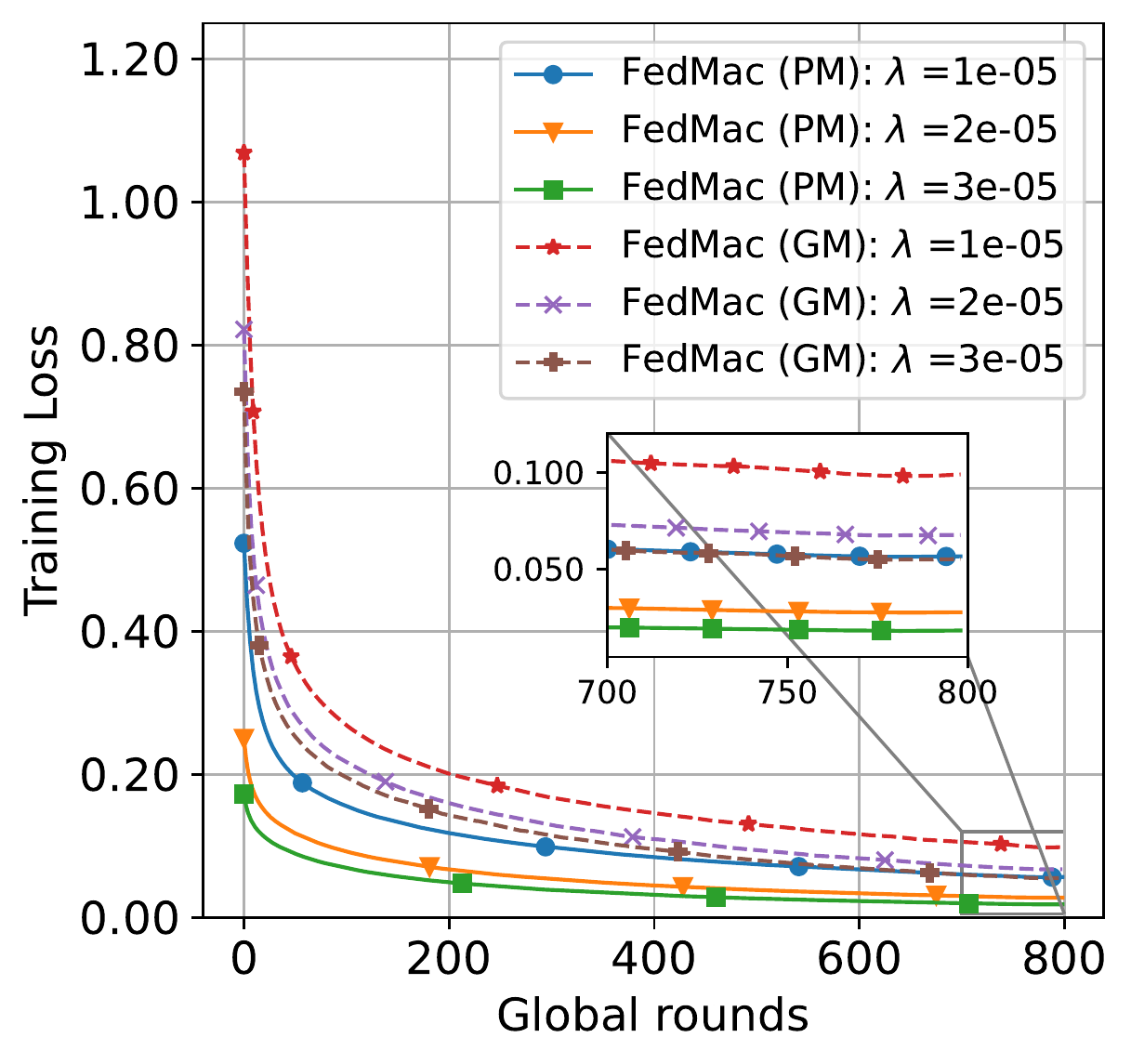}
	}
		
	\caption{\textcolor{black}{Effect of hyperparameters on the convergence rate of \texttt{FedMac} algorithm on MNIST dataset. \textbf{Top:} Test accuracy and training loss with different $R$. \textbf{Middle:} Test accuracy and training loss with different $\beta$. \textbf{Bottom:} Test accuracy and training loss with different $\lambda$.}}
	\label{fig_change_hyperparameters}
\end{figure}

\begin{figure}[!ht]
	\centering
	
	\subfloat{
	\includegraphics[width=2.5in]{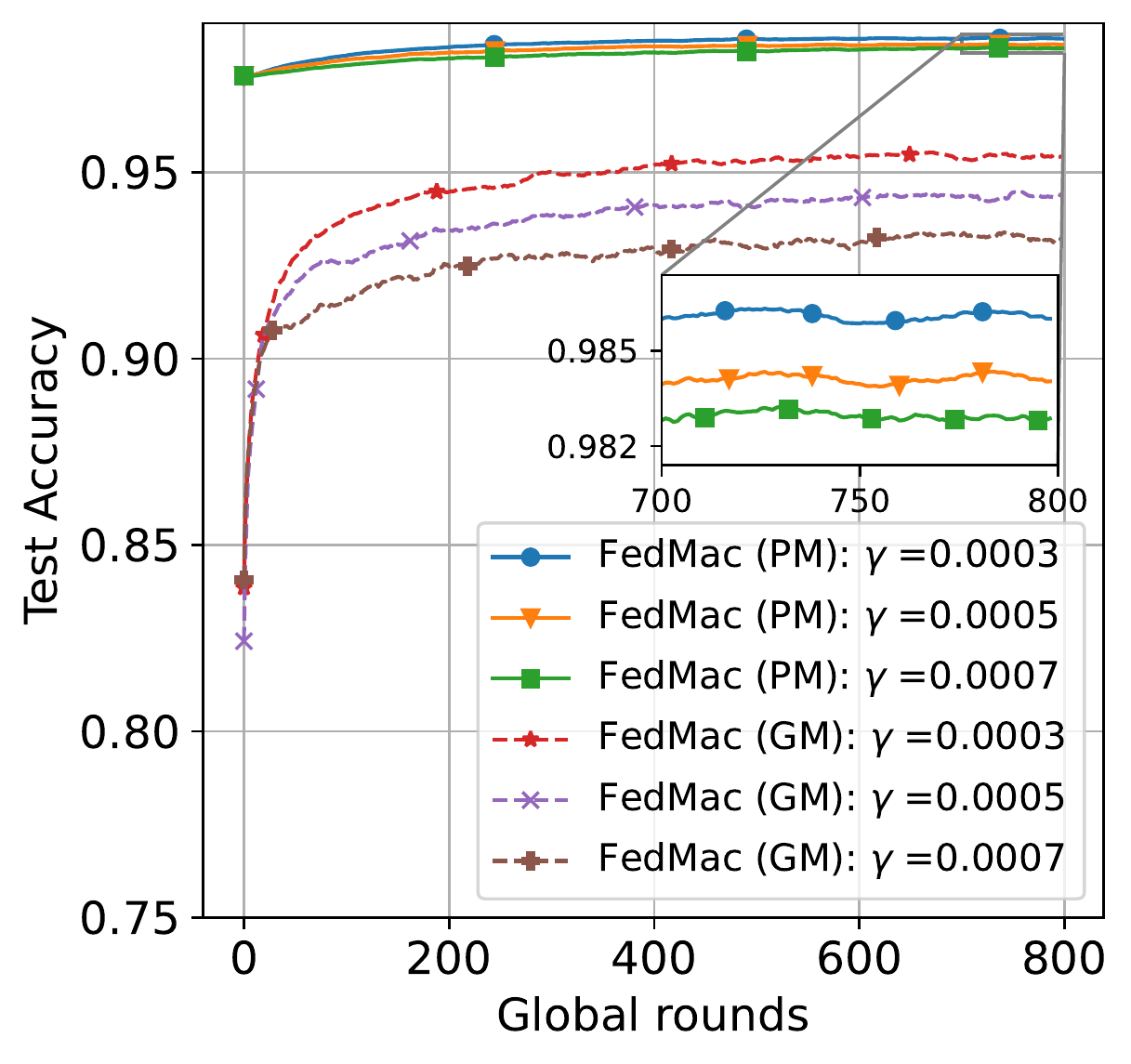}
	}
	\subfloat{
	\includegraphics[width=2.5in]{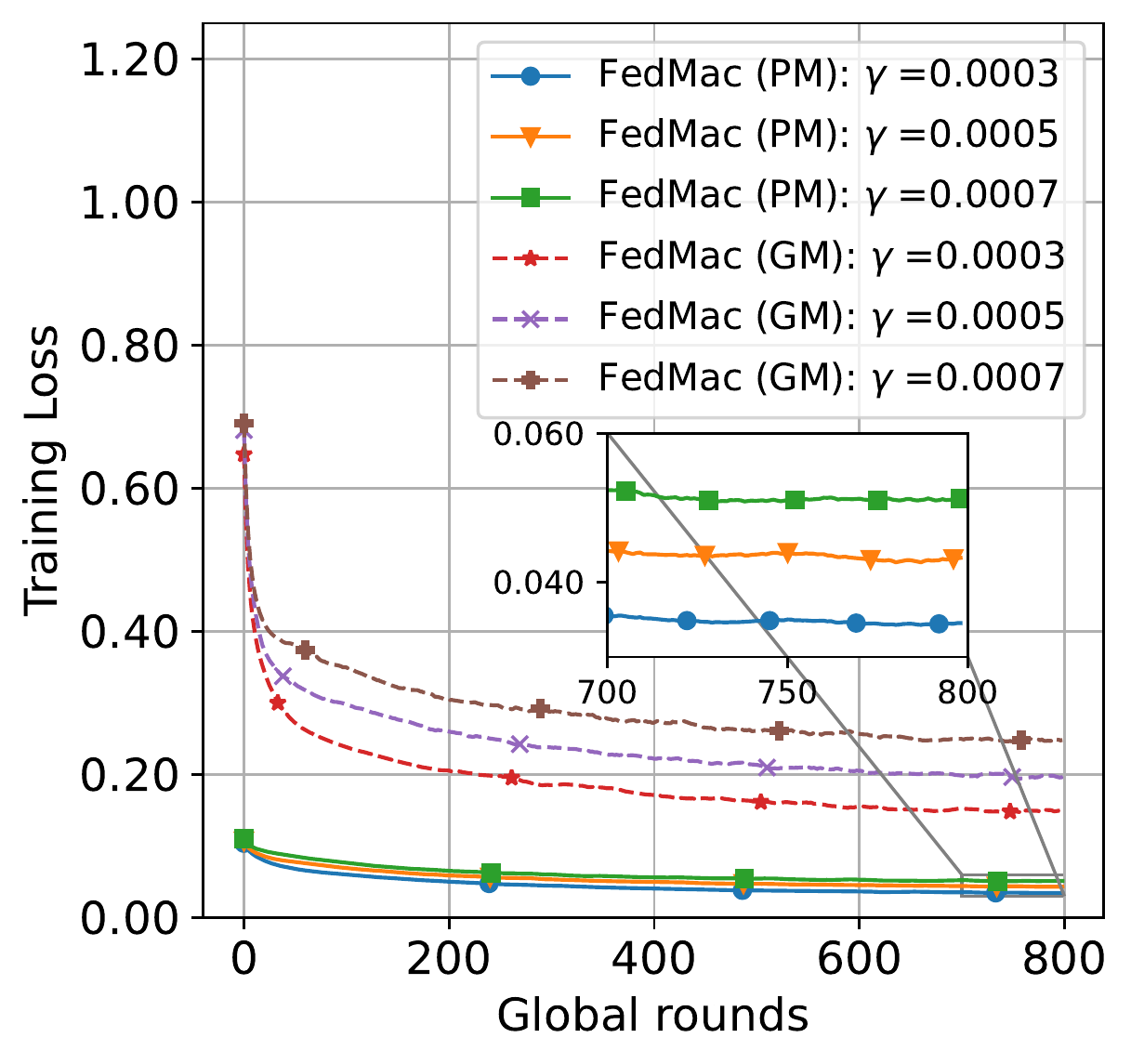}
	}

    \subfloat{
	\includegraphics[width=2.5in]{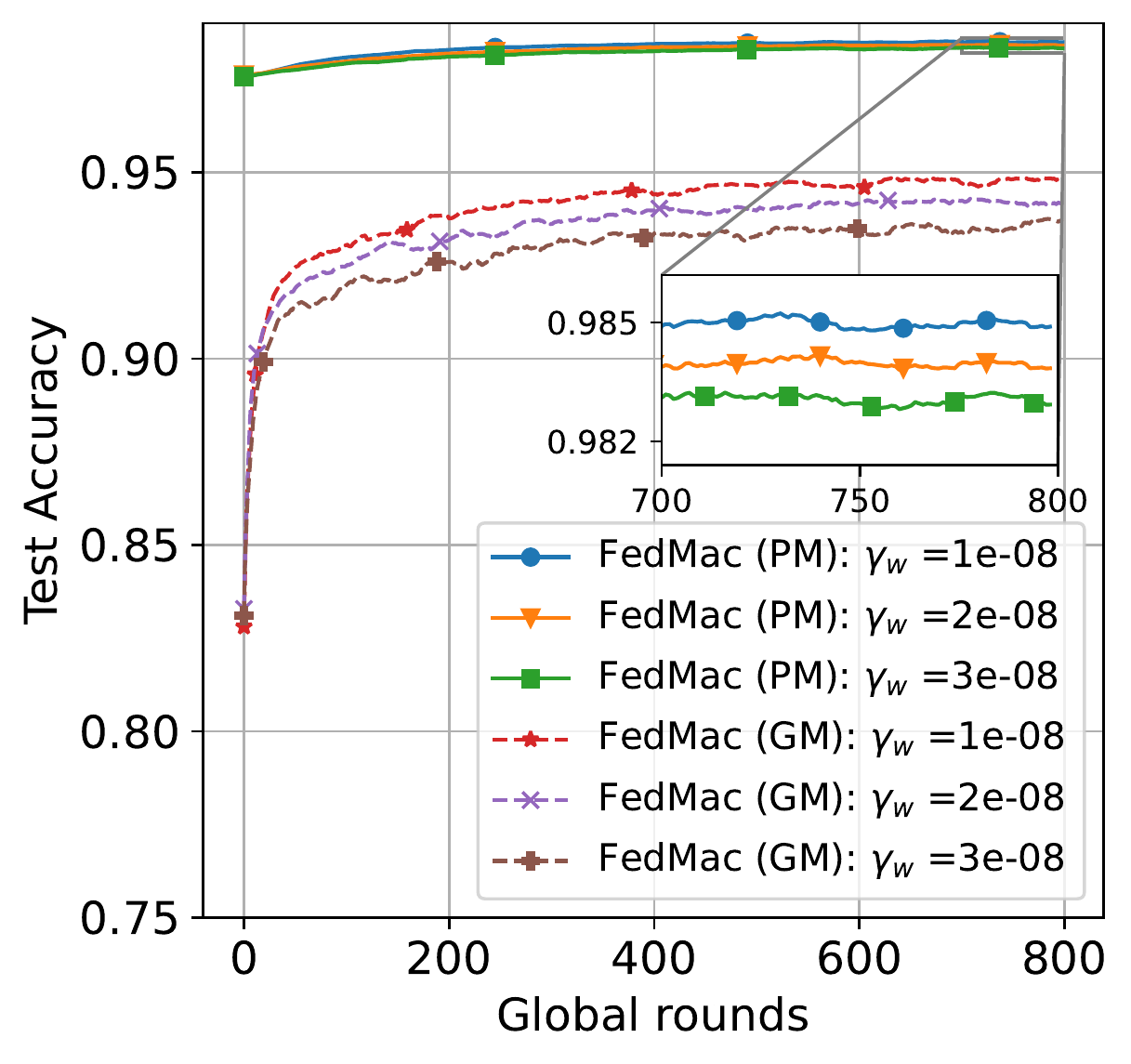}
	}
	\subfloat{
	\includegraphics[width=2.5in]{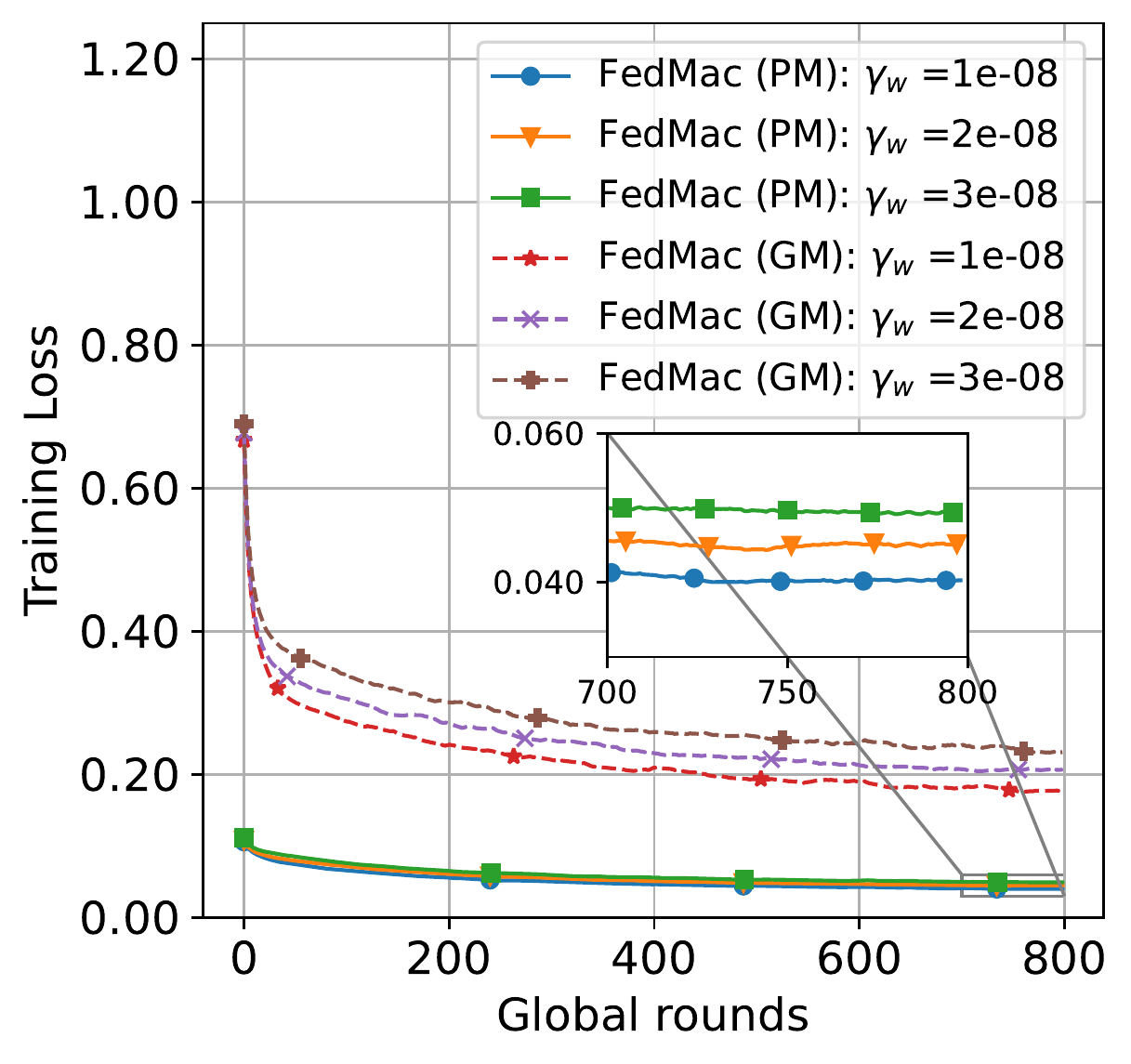}
	}
	\caption{\textcolor{black}{Effect of hyperparameters on the convergence rate of \texttt{FedMac} algorithm on MNIST dataset. \textbf{Top:} Test accuracy and training loss with different $\gamma$. \textbf{Bottom:} Test accuracy and training loss with different $\gamma_w$.}}
	\label{fig_change_hyp_gam_gw}
\end{figure}

{\color{black}
We first empirically study the effect of different hyperparameters in {\texttt{FedMac}} on MNIST dataset, where `GM' and `PM' denote the global model and personalized model respectively.

\textbf{Effects of $R$:} In {\texttt{FedMac}} algorithm, $R$ denotes the local epochs. An appropriately large $R$ allows the algorithm to converge faster but requires more computations at local clients, while a small $R$ requires more communication rounds between the server and clients. According to the top part of Figure~\ref{fig_change_hyperparameters}, where we fix $\eta=3000$, $\beta=1$, $S=10$, $|\mathcal{D}|=20$, $\lambda=0.0001$, $\gamma=\gamma_w=0$, we set $R=20$ for the remaining experiments to balance this trade-off.

\textbf{Effects of $\beta$:} The middle part of Figure~\ref{fig_change_hyperparameters} shows the test accuracy and training loss of {\texttt{FedMac}} algorithm with different values of $\beta$, where we set $\eta=3000$, $R=20$, $S=10$, $|\mathcal{D}|=20$, $\lambda=0.0001$, $\gamma=\gamma_w=0$. We can see that increasing $\beta$ can improve the test accuracy of the global model and weaken the performance of the personalized model. For balance, we set $\beta=1$ for the remaining experiments.

\textbf{Effects of  $\lambda$:} According to the bottom part of Figure~\ref{fig_change_hyperparameters}, where we fix $\eta=3000$, $\beta=1$, $S=10$, $|\mathcal{D}|=20$, $R=20$, $\gamma=\gamma_w=0$, properly increasing $\lambda$ can effectively improve the test accuracy and convergence rate for \texttt{FedMac}. And we find that an oversize $\lambda$ may cause gradient explosion. 

\textbf{Effects of $\gamma$:} The top part of Figure~\ref{fig_change_hyp_gam_gw} shows the relationship of $\gamma$ with the model accuracy and the convergence speed, where we fix $\eta=3000$, $\beta=1$, $S=10$, $|\mathcal{D}|=20$, $R=20$, $\lambda=0.0001$, $\gamma_w=0$. 
When $\gamma=0.0003$, $0.0005$ and $0.0007$, the model sparsity of \texttt{FedMac} are $0.81$, $0.78$ and $0.58$ respectively. 
We can see that increasing $\gamma$ reduces the model sparsity (rate of non-zero parameters) but decreases the model accuracy as well.

\textbf{Effects of $\gamma_w$:} The bottom part of Figure~\ref{fig_change_hyp_gam_gw} shows the relationship of $\gamma_w$ with the model accuracy and the convergence speed, where we fix $\eta=3000$, $\beta=1$, $S=10$, $|\mathcal{D}|=20$, $R=20$, $\lambda=0.0001$, $\gamma=0.0003$. 
When $\gamma_w=1\times10^{-8}$, $2\times10^{-8}$ and $3\times10^{-8}$, the model sparsity of \texttt{FedMac} are $0.82$, $0.72$ and $0.62$ respectively. 
We can see that increasing $\gamma_w$ reduces the rate of non-zero parameters but decreases the model accuracy as well.
}

\begin{figure}[!t]
	\centering
	\subfloat{
	\includegraphics[width=2.5in]{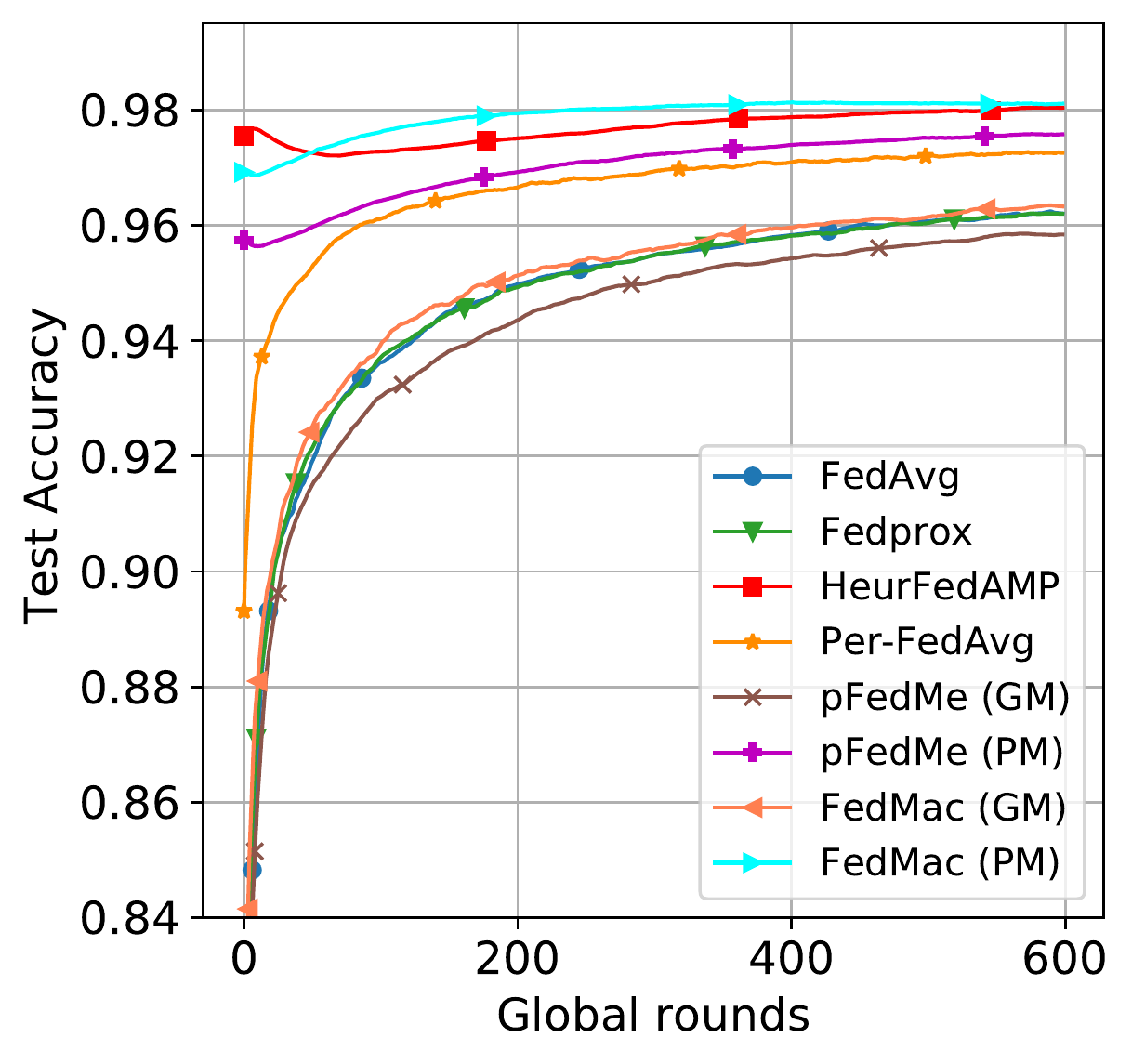}
	}
	\subfloat{
	\includegraphics[width=2.5in]{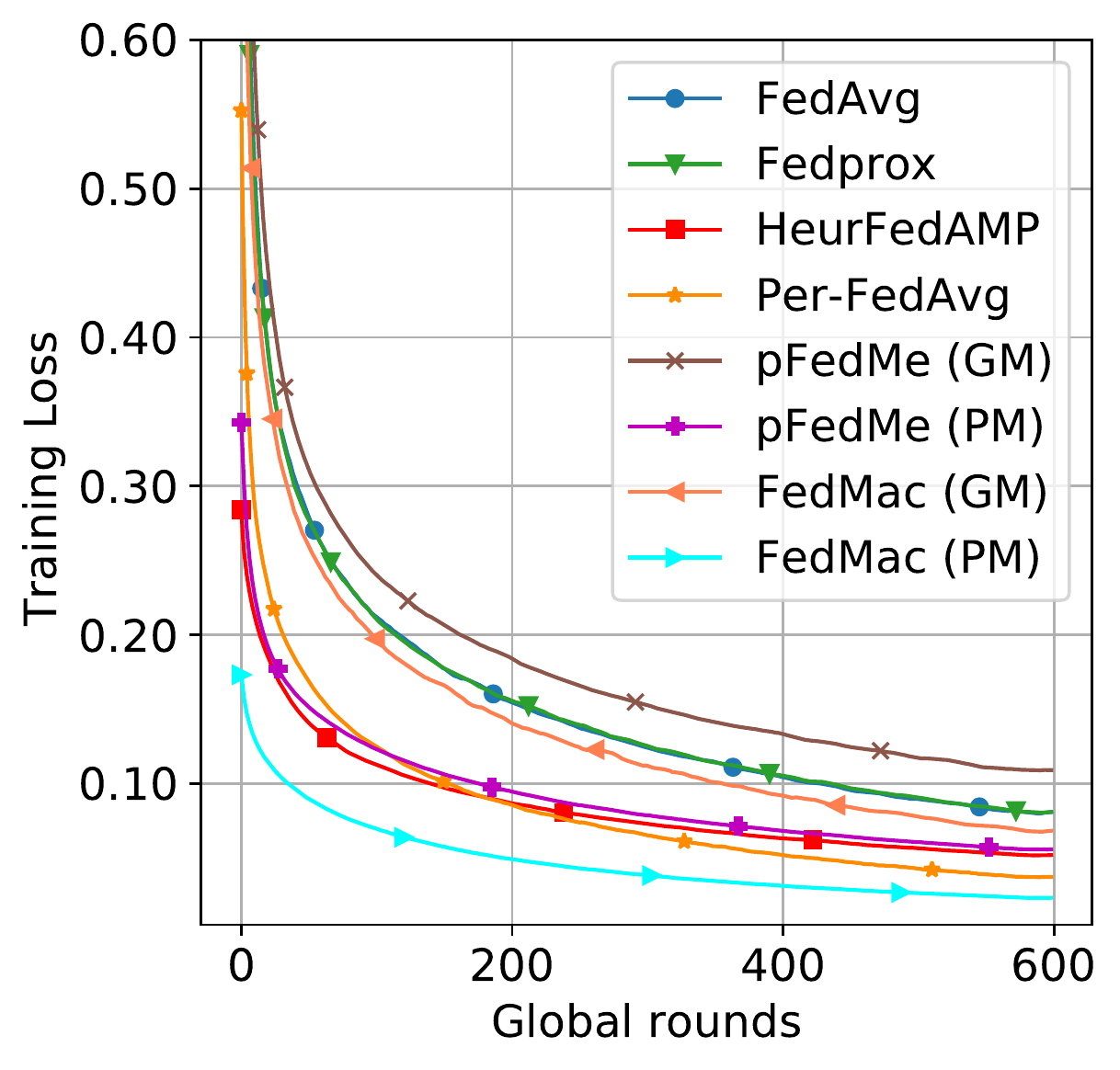}
	}
	\caption{Comparison results of the convergence rate of different algorithms on the MNIST dataset. \textbf{Left:} Test accuracy. \textbf{Right:} Training loss.}
	\label{fig_mnist_results}
\end{figure}

\begin{table*}[htbp]
\caption{Fine-tune results on MNIST, FMNIST, CIFAR-100, and Synthetic datasets. Best results are bolded.}
\label{fine-tune-table}
  \centering
  {\color{black}
  \scalebox{0.85}{
  \begin{tabular}{cccccccccccc}
    \toprule
    \multirow{2}{*}{Dataset}  & FedAvg & Fedprox & HeurFedAMP &Per-FedAvg &\multicolumn{2}{c}{pFedMe}  & \multicolumn{2}{c}{FedMac}  \\
    \cmidrule(r){2-2}
    \cmidrule(r){3-3}
    \cmidrule(r){4-4}
    \cmidrule(r){5-5}
    \cmidrule(r){6-7}
    \cmidrule(r){8-9}
    & GM & GM & PM & PM & GM & PM & GM & PM  \\
    \midrule
        {MNIST}         
        & 96.86$\pm 0.03$     
        & 96.86$\pm 0.04$     
        & 98.44$\pm 0.01$     
        & 98.15$\pm 0.05$     
        & 96.87$\pm 0.02$  & 98.92$\pm 0.01$     
        & \textbf{96.90$\pm$0.03}  & \textbf{98.95$\pm$0.01 }\\  
        {FMNIST}        
        & 85.69$\pm 0.12$     
        & 85.66$\pm 0.05$     
        & 99.18$\pm 0.01$     
        & 99.31$\pm 0.02$     
        & 84.96$\pm 0.20$  & 99.23$\pm 0.01$     
        & \textbf{85.80$\pm$0.17}  & \textbf{99.37$\pm$0.01}  \\ 
        {CIFAR-100}     
        & 85.77$\pm$1.73     
        & 86.02$\pm$0.71     
        & 90.40$\pm$0.75     
        & 89.78$\pm$0.38     
        & 84.95$\pm$1.43  & 90.14$\pm$0.11     
        & \textbf{86.57$\pm$0.27}  & \textbf{90.90$\pm$0.12}   \\   
        {Synthetic}     
        & 84.63$\pm$0.16     
        & 84.55$\pm$0.16     
        & 84.27$\pm$0.07     
        & 88.70$\pm$0.18     
        & 83.56$\pm$0.29  & 87.50$\pm$0.09     
        & \textbf{84.89$\pm$0.40}  & \textbf{89.06$\pm$0.02}   \\   
    \bottomrule
  \end{tabular}}
}
\end{table*}

\begin{table*}[htbp]
\caption{\textcolor{black}{Results for the same hyperparameter settings on MNIST, FMNIST, and Synthetic datasets.}}
\label{strong-result-table}
  \centering
  {\color{black}
  \scalebox{0.85}{
  \begin{tabular}{cccccccccccc}
    \toprule
    \multirow{2}{*}{Dataset}  & FedAvg & Fedprox & HeurFedAMP &Per-FedAvg &\multicolumn{2}{c}{pFedMe}  & \multicolumn{2}{c}{FedMac}  \\
    \cmidrule(r){2-2}
    \cmidrule(r){3-3}
    \cmidrule(r){4-4}
    \cmidrule(r){5-5}
    \cmidrule(r){6-7}
    \cmidrule(r){8-9}
    & GM & GM & PM & PM & GM & PM & GM & PM  \\
    \midrule
        {MNIST}         
        & 92.39$\pm 0.01$     
        & 92.40$\pm 0.03$     
        & 96.43$\pm 0.03$     
        & 93.05$\pm 0.03$     
        & 92.35$\pm 0.01$  & 95.70$\pm 0.07$     
        & \textbf{92.41$\pm$0.01}  & \textbf{96.64$\pm$0.05 } \\ 
        {FMNIST}        
        & \textbf{84.40$\pm$0.09}     
        & 84.35$\pm 0.03$     
        & 98.56$\pm 0.03$     
        & 98.38$\pm 0.01$     
        & 84.26$\pm 0.08$  & 98.88$\pm 0.01$     
        & 84.39$\pm$0.06   & \textbf{99.09$\pm$0.01} \\ 
        {Synthetic}     
        & 77.91$\pm$0.01     
        & 77.83$\pm$0.15     
        & 83.09$\pm$0.02     
        & 83.78$\pm$0.07     
        & 77.27$\pm$0.08  & 83.68$\pm$0.05     
        & \textbf{78.05$\pm$0.17}  & \textbf{84.40$\pm$0.02} \\ 
    \bottomrule
  \end{tabular}}
}
\end{table*}

\subsection{Performance Comparison Results}

We first compare the performance of our {\texttt{FedMac}} with other methods under non-sparse conditions to show its advantages, i.e., we set $\gamma=\gamma_w=0$ for  {\texttt{FedMac}}. Table~\ref{fine-tune-table} shows the fine-tune performances on different datasets and models. We test these algorithms through the same settings with $|\mathcal{D}|=20$ and $T=800$, and appropriately fine-tune other hyperparameters to obtain the highest mean testing accuracy in all communication rounds of training. We run each experiment at least 3 times to obtain statistical reports. More detailed results on hyperparameters are listed in the appendix. We can see that the personalized model (PM) of {\texttt{FedMac}} outperforms other models in all settings, while the global model (GM) of {\texttt{FedMac}} outperforms all other global models.

\begin{table}[ht] 
\centering
{
\color{black}
\caption{\textcolor{black}{Results on for larger dataset and bigger model task.}}
  \label{big-model-table}
  \scalebox{1}{
  \begin{tabular}{cccccc}
    \toprule
     \multirow{2}{*}{Method}  
    & \multicolumn{2}{c}{$S/N=5/10$} 
    & \multicolumn{2}{c}{$S/N=10/10$}\\
    \cmidrule(r){2-3}
    \cmidrule(r){4-5}
    & GM
    & PM
    & GM
    & PM\\
    \midrule
        FedAvg    
       & 54.68$\pm$0.26 & - 
       & 57.35$\pm$0.07 & - \\
        Fedprox    
       & 68.23$\pm$0.89 & - 
       & 68.72$\pm$0.19 & - \\
        HeurFedAMP      
       & -  & 68.55$\pm$0.13
       & -  & 71.35$\pm$0.32\\
        Per-FedAvg     
       & -  & {72.06$\pm$0.16}
       & -  & {71.69$\pm$0.31}\\
        pFedMe     
       & 68.41$\pm$0.13  & 70.48$\pm$0.06
       & 68.86$\pm$0.37  & 69.64$\pm$0.25\\
        FedMac       
       & \textbf{68.97$\pm$0.27}  & \textbf{73.52$\pm$0.33}
       & \textbf{69.70$\pm$0.05}  & \textbf{73.33$\pm$0.09}\\
    \bottomrule
  \end{tabular}}
}  
\end{table}

Figure~\ref{fig_mnist_results} shows the convergence rate of our {\texttt{FedMac}} algorithm and other algorithms on the MNIST dataset. The accuracy of the personalized model of {\texttt{FedMac}} is 1.79\%, 0.55\%, 2.29\%, 0.75\%, 0.13\%, 1.96\%, and 1.92\% higher than that of the global model of {\texttt{FedMac}}, the personalized model of pFedMe, the global model of pFedMe, Per-FedAvg, HeurFedAMP, Fedprox and FedAvg, respectively. In these experiments, we set $R=20$, $S=10$, $|\mathcal{D}|=20$, $\beta=1$ for all algorithms. For {\texttt{FedMac}}, we use small $\lambda = 0.0001$ because  Theorem~\ref{UpperBound} shows that a small $\zeta = \lambda/\gamma$ can reduce the upper bound of the required training data. For pFedMe, we set $\lambda = 15$, which is a recommended value in~\cite{t2020personalized}. To balance the effects of different $\lambda$ on different convergence speeds, we set $\eta=0.02$ for pFedMe and $\eta=3000$ for {\texttt{FedMac}} to make $\eta \times \lambda$ equal. For other hyperparameters in other algorithms, we try to set the same value for a fair comparison. See the appendix for detailed settings.

{
\color{black}
Table~\ref{strong-result-table} shows the performances for $\mu$-strong convex setting, in which a multinomial logistic regression model (MLR) is considered with the softmax activation and cross-entropy loss following the convex setting in~\cite{t2020personalized}. On MNIST, MLR task, our method achieves the best accuracy in both global model (92.41\%) and personalized model (96.64\%). On Synthetic, MLR task, our method also achieves the highest accuracy in personalized model (99.09\%), with global model performance (84.39\%) similar to FedAvg (84.40\%). On Synthetic, MLR task, our method still achieves the best results as shown in Table~\ref{strong-result-table}.
}
\begin{figure}[tp]
	\centering
	\includegraphics[height=2.5in]{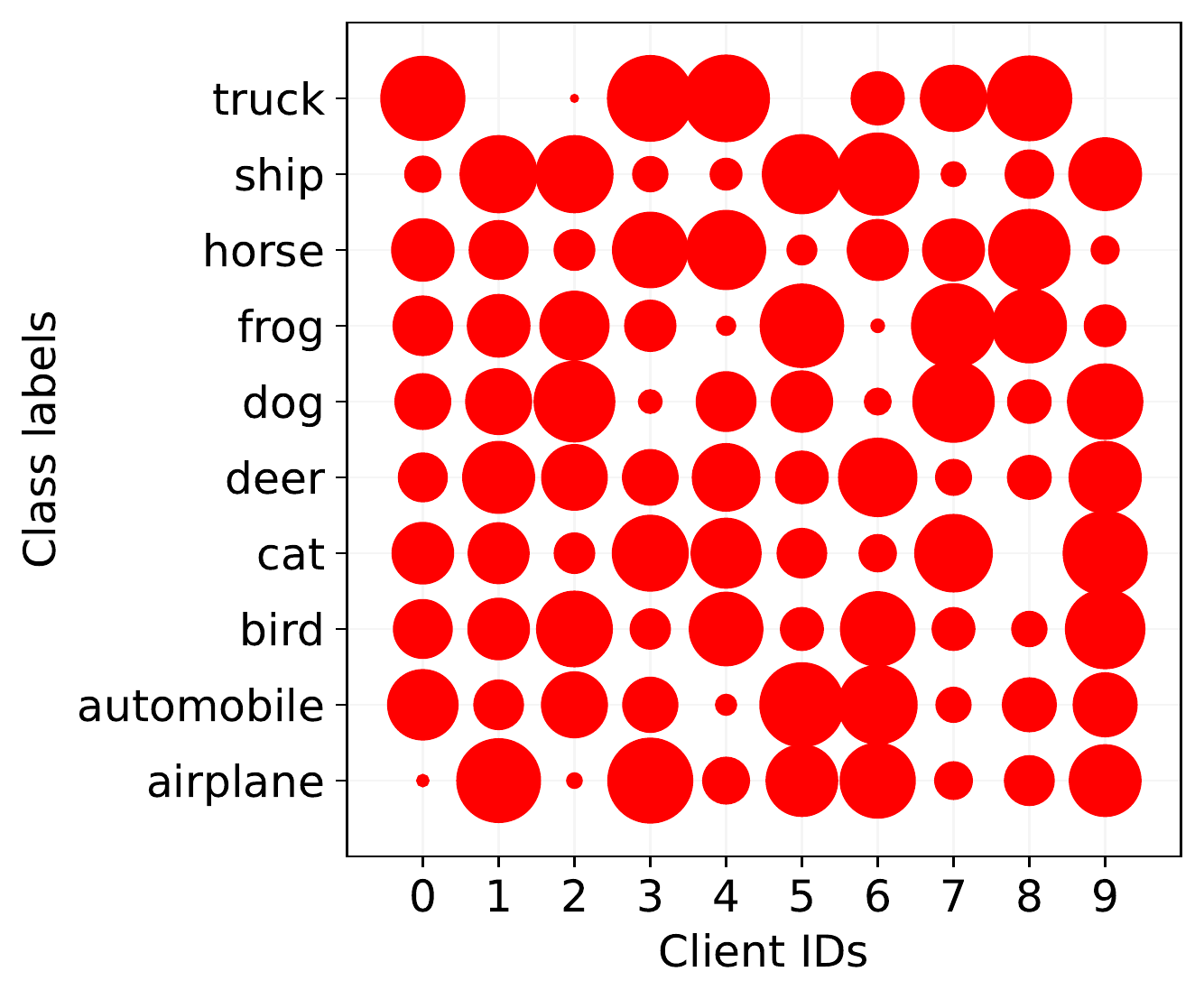}
	\caption{\textcolor{black}{Illustration of samples distribution for different Dirichlet distribution with $\alpha=1$ on CINIC-10 dataset.}}
	\label{fig_noniid_distribution}
\end{figure}

{\color{black}
Table~\ref{big-model-table} shows the results on CINIC-10, WRN task for two different selection strategies. we set $|\mathcal{D}|=64,~ R=50,~T=300$, and $\beta=1$ for all algorithms and set $\{\eta=0.06,~\lambda=15\}$ for pFedMe and $\{\eta=9000,~\lambda=0.0001\}$ for {\texttt{FedMac}} to make $\eta \times \lambda$ equal. Specifically, we find that $\lambda$ in FedAMP is sensitive to the model, while the recommended setting $\lambda=1$ in \cite{huang2021personalized} does not work for WRN. Thus, we fine-tune $\lambda$ around 1 and use $\lambda=0.01$ for FedAMP to obtain a better result. 
For the non-IID setting, we use the Dirichlet distribution as in~\cite{NEURIPS2020_18df51b9} to create disjoint client training data. The parameter $\alpha$ controls the degree of non-IID, we use $\alpha=1$ and show the sample distribution in Figure~\ref{fig_noniid_distribution}.
According to results in Table~\ref{big-model-table}, we can see FedAvg performs poorly on the CINIC-10 dataset with non-IID setting compared with personalized FL methods. Our proposed method achieves the best accuracy in both global and personalized models.
}

{
\color{black}
\subsection{Communication Discussion}

{\color{black}
First, we show the performance advantages of our {\texttt{FedMac}} over the $\ell_2$-norm-based personalized methods under sparse conditions. To control the variables, we set $\gamma_w=0$ in this experiment, that is, we only analyze the difference in  sparsity caused by different optimization functions in (4). For comparison, we obtain modified FedProx, Per-FedAvg, and pFedMe algorithms by adding approximate $\ell_1$-norm constraints to the loss function. Table~\ref{sparse-com-table} shows that our {\texttt{FedMac}} algorithm has significantly better performance than the $\ell_2$-norm based personalized methods under sparse conditions. 
}
\begin{table}[ht] 
\centering
{
\color{black}
\caption{\textcolor{black}{Results on for sparse setting.}}
  \label{sparse-com-table}
  \scalebox{1}{
  \begin{tabular}{cccccc}
    \toprule
     \multirow{1}{*}{Method}  
    & \multirow{1}{*}{Sparsity}
    & GM
    & PM\\
    \midrule
        Fedprox    
       & 74.38\% 
       & 93.56$\pm$0.05 & - \\
        Per-FedAvg     
       & 83.56\%
       & -  & {97.49$\pm$0.06}\\
        pFedMe     
       & 71.62\%
       & 93.29$\pm$0.06  & 97.40$\pm$0.07\\
        FedMac       
       & \textbf{49.50\%} 
       & \textbf{95.08$\pm$0.03}  & \textbf{98.33$\pm$0.06}\\
    \bottomrule
  \end{tabular}}
}  
\end{table}

\begin{figure}[ht]
	\centering
	
	\subfloat{
	\includegraphics[width=2.5in]{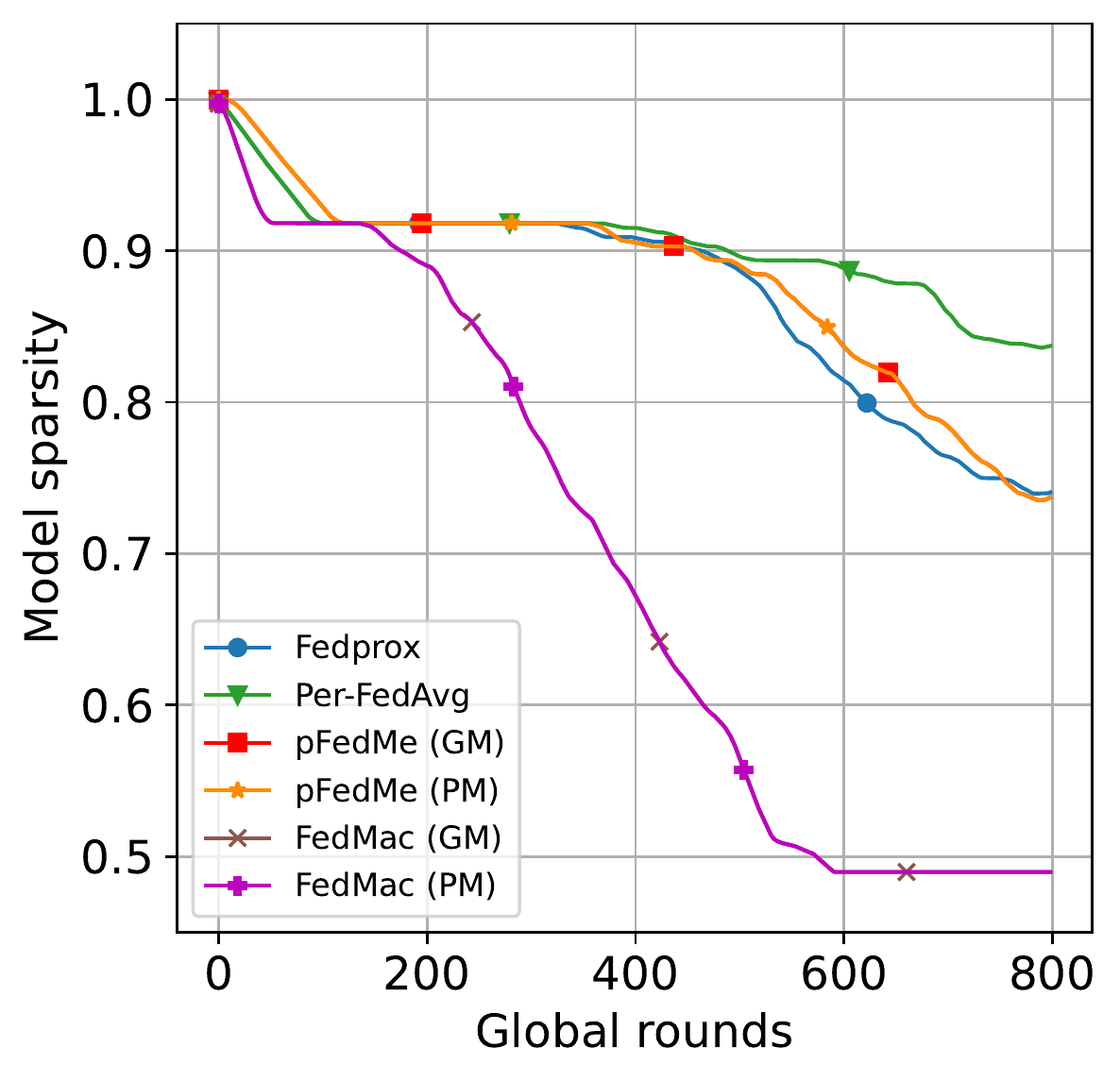}
	}
    \quad
	\subfloat{
	\includegraphics[width=2.5in]{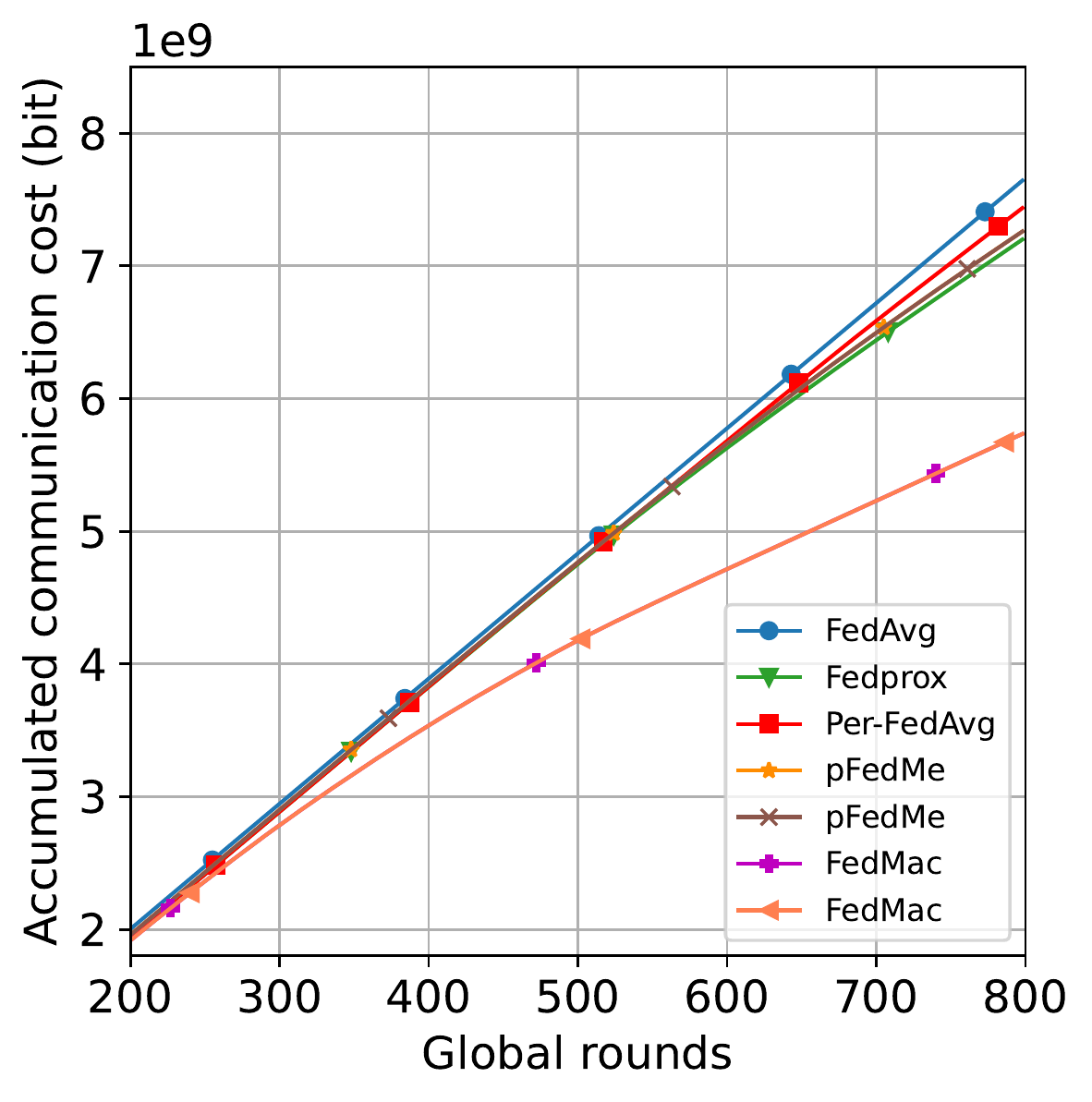}
	}
	
	\caption{\textcolor{black}{The model sparsity and communication cost for different algorithms.}}
	\label{fig_sparse_com}
\end{figure}

{\color{black}
The model sparsity and the accumulative communication cost against  the global rounds for experiments in Table~\ref{sparse-com-table} are presented in Figure~\ref{fig_sparse_com}. Sparsity is defined as the proportion of the number of non-zeros in the model, where we set 50\% sparsity as a lower bound to preserve performance. The accumulated communication cost is calculated based on a model with 79510 parameters. Non-zero parameters are quantized by 64 bits, while zero parameters are quantized by 1 bit.
And the {\texttt{FedMac}} algorithm needs to communicate an additional index matrix to represent the location of the zero value, which requires $79510\times1$ bits of communication cost. 
Therefore, for each round of iterative training, the communication cost required by the server and each client for the non-sparse model is $79510\times2\times64$ bits for upload and broadcast, while the communication cost required by the server and each client for the sparse model is $\text{sparsity}\times 79510\times2\times64 + (1-\text{sparsity})\times 79510\times2\times1 + 79510\times1 $ bits for upload and broadcast. With a 50\% sparsity setting, the communication cost is reduced by almost half.
}

}

\section{Conclusions}\label{sec_conclusion}

In this paper, we propose {\texttt{FedMac}} as a sparse personalized FL algorithm to solve the statistical diversity issue, which has better performance than the $\ell_2$-norm  based personalization methods under sparse conditions. Our approach makes use of an approximated $\ell_1$-norm and the correlation between the global model and client models in the loss function. Maximizing correlation decouples the personalized model optimization from the global model learning, which allows {\texttt{FedMac}} to optimize personalized models in parallel. Convergence analysis indicates that the sparse constraints in {\texttt{FedMac}} do not affect the convergence rate of the global model. Moreover, theoretical results show that {\texttt{FedMac}} performs better than the $\ell_2$-norm based personalization methods and the training data required is significantly reduced. Experimental results demonstrate that {\texttt{FedMac}} outperforms many advanced personalization methods under both sparse and non-sparse conditions.




\bibliographystyle{unsrt} 
\bibliography{references.bib}

\appendix

\subsection{Proof of Theorem~\ref{theorem_convergence_FedMac}}

\begin{proof}
We first prove part $(a)$, for any $\tilde\eta \leq \hat\eta \triangleq \frac{1}{(18+256\kappa_F)L_F} $ with $\beta \geq 1$ and $L_F\geq \mu_F$ we have
\begin{align}
    \tilde\eta \leq \min \left\{ \frac{\beta\sqrt{R}}{2L_F \sqrt{1+R}} , \frac{1}{\mu_F},    \frac{1}{(18+256\kappa_F)L_F}  \right\}.  \label{constraint_eta}
\end{align}

Then, we substitute Lemma~\ref{lemma_g-F}, Lemma~\ref{lemma_Fi-F} and Lemma~\ref{lemma_Fi-F_s} into Lemma~\ref{lemma_one_step} to obtain
\begin{align}
    &\mathbb{E}  \left[ \| \bm w^{t+1} - \bm w^{\star} \|^2_2 \right] \notag
    \overset{(a)}{\leq}&~   \mathbb{E}  \left[ \| \bm w^{t} - \bm w^{\star} \|^2_2 \right]  - \tilde\eta \left(  F(\bm w^t) -F(\bm w^{\star}) \right) + \tilde\eta M_1 +  \tilde\eta^2 M_2 ,  \notag
\end{align}
where $(a)$ is due to $\frac{N/S-1}{N-1} \leq 1$ and $3\tilde\eta + 1/\mu_F \leq 4/\mu_F$ by setting $\tilde\eta \leq 1/\mu_F$, $2 - 256 \tilde\eta L_F \kappa_F - 18\eta L_F \geq 1$  by setting $\tilde\eta \leq \frac{1}{(18+256\kappa_F)L_F}$ and define $\kappa_F = L_F/\mu_F$, $M_1=\frac{32 \sigma_F^2 + 40 \lambda^2\delta^2}{\mu_F}$ and $M_2=6\sigma_F^2 (\frac{N/S-1}{N-1})$.

Then we have
\begin{align}
     \mathbb{E} \left [F(\bm w^t) -F(\bm w^{\star})  \right] \leq \frac{1}{\tilde\eta} (\Delta_t - \Delta_{t+1})  + M_1 + \tilde\eta M_2 ,  \notag
\end{align}
where $\Delta_t \triangleq \mathbb{E} \left[  \| \bm w^t - \bm w^{\star}  \|^2_2 \right]$. By taking average over $T$, we have
\begin{align}
    \frac{1}{T} \sum_{t=0}^{T-1}  \mathbb{E} \left [F(\bm w^t) -F(\bm w^{\star})  \right] 
    \leq&~ \frac{\Delta_0}{\tilde\eta T} + M_1 + \tilde\eta M_2 \label{eq_delta0-T}.
\end{align}

We now consider two cases:
\begin{itemize}
    \item If 
    $\hat\eta \geq  \left(\frac{\Delta_0}{M_2 T}\right)^{\frac{1}{2}}$, we choose $\tilde\eta = \left(\frac{\Delta_0}{M_2 T}\right)^{\frac{1}{2}}$ 
    
    \item If 
    $\hat\eta <  \left(\frac{\Delta_0}{M_2 T}\right)^{\frac{1}{2}}$, we choose $\tilde\eta = \hat\eta$
\end{itemize}

Collecting the two cases above, we get a bound as follow:
\begin{align}
\label{theorem1-parta}
    \frac{1}{T} \sum_{t=0}^{T-1}  \mathbb{E} \left [F(\bm w^t) -F(\bm w^{\star})  \right] 
    \leq&~ \mathcal{O}    \left\{   \frac{\Delta_0}{\hat\eta T} + \frac{ \sigma_F^2 +  \lambda^2\delta^2}{\mu_F} + \frac{ (\sigma_F^2 \Delta_0(N/S-1))^{1/2}}{\sqrt{TN}}    \right\} ,   
\end{align}

We next prove part $(b)$. First we have

\begin{align}
    \frac{1}{N} \sum_{i=1}^N \mathbb{E} \left[  \|  \tilde{\bm \theta}^T_i(\bm w^T)   -  \bm w^{\star} \|^2_2  \right]   
    \overset{(a)}{\leq}&~ \frac{4}{N} \sum_{i=1}^N \mathbb{E} \bigg[   \frac{1}{\lambda^2} \| \nabla F_i (\bm w^T)  \|^2_2     +       \| \frac{\gamma_w}{\lambda}\nabla \phi_{\rho} (\bm w^T)  \|^2_2 \bigg] \notag\\
    &~ + 4\delta^2 + 4\Delta_T,  \label{eq_thetaT-w}
\end{align}
where $(a)$ is due to Jensen's inequality and Lemma~\ref{lemma_theta-theta}. Note that
\begin{align}
    \frac{1}{N} \sum_{i=1}^N\mathbb{E} \bigg[   \| \nabla F_i (\bm w^T)  \|^2_2 \bigg] 
    \overset{(a)}{\leq}&~ 2 L_F^2 \Delta_T +2 \sigma_F^2,  \label{eq_theta-w_1}
\end{align}
where $(a)$ is due to Jensen's inequality and $F_i$ is $L_F$-smooth, and
\begin{align}
    \mathbb{E}\left[\left\|\nabla\phi_{\rho}({\bm{w}^T})\right\|_2^2\right]
    &\leq \mathbb{E}\left[\sum_{i=1}^{d}\left( \frac{w^{T}_n}{\left|w^{T}_n\right|}\right)^2\right]  = d_s^2\label{eq_theta-w_2},
\end{align}
where 
$w^{T}_n$ is the $n$-th element of $\bm w^T$ and $d_s$ denotes the number of non-zero elements in $\bm w$.

Since $F$ is $\mu_F$-strongly convex,  we have
\begin{align}
    \Delta_T 
    \leq&~ \frac{2}{\mu_F} \frac{1}{T} \sum_{t=0}^{T-1}  \mathbb{E} \left [F(\bm w^t) -F(\bm w^{\star})  \right]  \label{eq_theta-w_4}.
\end{align}

Substituting \eqref{theorem1-parta}, \eqref{eq_theta-w_1}, \eqref{eq_theta-w_2} and \eqref{eq_theta-w_4} into \eqref{eq_thetaT-w}, we complete the proof.
\end{proof}

\subsection{Proof of Theorem~\ref{theorem_Base}}

\begin{proof} Since  $\hat{\bm \theta}$ is the solution of
\begin{align}
      \min _{\bm \theta \in \mathbb{R}^{d} }   \norm{\bm y - \bm X \bm \theta}_2^2 + \gamma f(\bm \theta), \notag
\end{align}
we have
\begin{equation} 
	\norm{\vy-\vX\hat{\bm \theta}}_2^2 + \gamma f(\hat{\bm \theta}) \le \norm{\vy-\vX\bm \theta^\star}_2^2+ \gamma f(\bm \theta^\star) = \gamma f(\bm \theta^\star). \notag
\end{equation}	

Let $\vh = \hat{\bm \theta} - \bm \theta^\star$.  We obtain
\begin{equation}  \label{neq: solution_condition2} 
	\norm{\vX\vh}_2^2
	\le	  \gamma f(\bm \theta^\star)- \gamma f(\hat{\bm \theta}).
\end{equation}	

Using $\norm{\vX\vh}_2^2>0$ yields $f(\hat{\bm \theta}) \le	 f(\bm \theta^\star)$.

So $\vh$ is in the error set $\mathcal{E}_f$ and hence is in the set $\mathcal{C}_f$, i.e. $\vh \in \mathcal{C}_f$. Let $\bar{\vh}=\vh/\norm{\vh}_2 \in \mathcal{C}_f \cap \S^{N_I-1}$ with $\S^{N_I-1}$ being an $N_I$-dimensional sphere.
By using Lemma \ref{lm:MatrixDeviationInequality} as following, with probability at least $1- 2\exp(-t^2)$.
\begin{lemma}[Matrix deviation inequality, {\cite[Theorem 3]{liaw2016simple}}] \label{lm:MatrixDeviationInequality}
	Let $\vX$ be an ${N_D \times N_I}$ random matrix whose rows $\{\vX_{i}\}_{i=1}^{N_D}$ are independent, centered, isotropic and sub-Gaussian random vectors. For any bounded subset $\mathcal{D} \subset \R^{N_I}$ and $t \ge 0$,  the following  inequality holds with probability at least $1-2\exp(-t^2)$
	$$
	\sup \limits_{\bm \theta \in \mathcal{D}} \left| \norm{\vX \bm \theta}_2 - \sqrt{N_D} \norm{\bm \theta}_2 \right| \le CK^2 [\xi (\mathcal{D})+t\cdot \text{\emph{rad}}(\mathcal{D})],
	$$
	where $ \text{\emph{rad}}(\mathcal{D})=\sup_{\bm \theta \in \mathcal{D}} \norm{\bm \theta}_2$ and $K=\max_i \norm{\vX_{i}}_{\psi_2}$.
\end{lemma}

Then, we have 
\begin{align}\label{Restricted_Eigenvulue}
	\sqrt{N_D} - \inf \limits_{\bar{\vh}} \norm{\vX \bar{\vh}}_2 
	\leq \sup \limits_{\bar{\vh}} \left| \norm{\vX \bar{\vh}}_2 - \sqrt{N_D} \right| 
	\leq C' K^2 [\xi(\mathcal{C}_f \cap \S^{N_I-1}) + t],
\end{align}
where $\bar{\vh} \in \mathcal{C}_f \cap \S^{N_I-1}$ and the result follows from that $\norm{\bar{\vh}}_2=1$ and $\text{rad}(\mathcal{C}_f\cap \S^{N_I-1})=1$. Let $t = \xi(\mathcal{C}_f \cap \S^{N_I-1})$. Then if $\sqrt{N_D} \ge CK^2 \xi(\mathcal{C}_f \cap \S^{N_I-1})+ \epsilon$, with probability at least $1- 2\exp(-\xi^2(\mathcal{C}_f \cap \S^{N_I-1}))$ we get
\begin{align}
	\inf \limits_{\bar{\vh} \in \mathcal{C}_f \cap \S^{N_I-1}}\norm{\vX \bar{\vh}}_2\geq  \sqrt{N_D}-CK^2 \xi(\mathcal{C}_f \cap \S^{N_I-1})\geq \epsilon , \notag
\end{align}
where $C = 2C'$. So we have
\begin{equation}\label{eq: MatrixDeviationInq_2}
	\|\vX\vh\|_2 = \norm{\hat{\bm \theta}-\bm \theta^\star}_2 \norm{\vX \frac{\vh}{\norm{\vh}_2}}_2 \geq \epsilon \norm{\hat{\bm \theta}-\bm \theta^\star}_2.
\end{equation}

On the other hand,  combining \eqref{neq: solution_condition2}  and the fact that $|f(\bm \theta^\star) - f(\hat{\bm \theta})|\le \alpha_f ||\hat{\bm \theta}-\bm \theta^\star||_2$, we have
\begin{equation}  \label{neq: solution_condition3} 
	\norm{\vX\vh}_2^2
	\le	  \gamma f(\bm \theta^\star)- \gamma f(\hat{\bm \theta})
	\le \gamma \alpha_f \norm{\hat{\bm \theta}-\bm \theta^\star}_2.
\end{equation}

If  $\gamma \leq 1 / \alpha_f$, combining \eqref{eq: MatrixDeviationInq_2} and \eqref{neq: solution_condition3} obtains
\begin{equation}  \label{neq: solution_condition4} 
	\norm{\hat{\bm \theta}-\bm \theta^\star}_2 \le \frac{\gamma\alpha_f}{\epsilon^2} \le \frac{1}{\epsilon^2},
\end{equation}	
with the desired probability. Then we complete the proof.
\end{proof}

\subsection{Proof of Theorem~\ref{UpperBound}}


\begin{proof}

To start with, based on Lemma~\ref{lemma2}:
\begin{lemma} \label{lemma2}The Gaussian complexity $\xi (\mathcal{C}_f \cap \S^{N_I-1})$ satisfies 
	\begin{equation}
		\xi (\mathcal{C}_f \cap \S^{N_I-1})\le 2 \eta(\gamma \, \partial f(\bm \theta^\star))+1.
	\end{equation}
\end{lemma}

We have
\begin{equation}
		\xi (\mathcal{C}_f \cap \S^{N_I-1})\le 2 \eta(\gamma \, \partial f(\bm \theta^\star))+1.
	\end{equation}

Next, we only need to compare $\eta(\gamma \, \partial f_1(\bm \theta^\star))$ and $\eta(\gamma \, \partial f_2(\bm \theta^\star))$ for the comparison of $f_1(\bm \theta)$ and $f_2(\bm \theta)$. In order to use the smallest number of measurements, we choose 
\begin{align}
	\gamma_1^\star&= \arg \min_\gamma \eta^2(\gamma \, \partial f_1(\bm \theta^\star)),\\
	\gamma_2^\star&= \arg \min_\gamma \eta^2(\gamma \, \partial f_2(\bm \theta^\star)).
\end{align}
Then we will give the upper bound for $\eta^2(\gamma_1^\star \, \partial f_1(\bm \theta^\star))$ and $\eta^2(\gamma_2^\star \, \partial f_2(\bm \theta^\star))$ by following the method developed in \cite[Theorem 14]{mota2017compressed}. Recall that, $I=\{n:  \theta_n^\star \neq 0\},~J=\{n:  \theta_n^\star \neq w_n\},~q=|I \cup J|$ and we further introduce the following definitions
$$I^c=\{n:  \theta_n^\star = 0\},~ J^c=\{n:  \theta_i^\star = w_n\}, $$
$$K_\zeta^{\neq}=\{ n \in I^cJ: |w_n |>1/{\zeta}\},$$
$$K_\zeta^{\neq}=\{ n \in I^cJ: |w_n |>1/{\zeta}\},$$
$$\varphi(\theta)=\frac{1}{\sqrt{2 \pi}} \exp\left(-\frac{\theta^2}{2}\right),$$ 
where  $I^cJ$ means $I^c \cap J$.
Let $s=|I|$ and $\overline{w}=|w_k|$, where $k=\arg \min_{i \in I^cJ}\left| | w_i |-\frac{1}{\zeta}\right|$. Here, $\zeta=\zeta_1$ for $f_1(\bm \theta)$ and $\zeta=\zeta_2$ for $f_2(\bm \theta)$.

Recall that
$$v_{{\zeta_1} } = \sum_{n \in I}(\text{sign}(\theta_n^\star)-{\zeta_1} w_n)^2+\sum_{n \in K_{\zeta_1}^{\neq}} ({\zeta_1} |w_n|-1)^2,$$
$$v_{\zeta_2}= \sum_{n \in I}(\text{sign}(\theta_n^\star)+\zeta_2 (\theta_n^\star-w_n))^2+\sum_{n \in K_{\zeta_2}^{\neq}} (\zeta_2 |w_n|-1)^2,
$$
where $\sign(\theta)$ denotes the sign function, i.e.,
\begin{equation*}
	\sign(\theta)=
	\left\{
	{\begin{array}{rl}
			1, & x>0,  \\
			0, & x=0,  \\
			-1,& x<0.
	\end{array} }
	\right.
\end{equation*}

Then we have the following results.

\begin{lemma} \label{lemma3}
 Let $\bm \theta^\star \in \R^{N_I}$ be an $s$-sparse vector and $\vw \in \R^{N_I}$ be its similar information. Let $f_1$ denote the convex function $ f_1(\bm \theta):=\norm{\bm \theta}_1 - \zeta_1 \ip{\vw}{\bm \theta}$.
	

	\begin{equation}\label{condition1}
	\textrm{If~}
	\frac{q\!-\!s}{N_I\!-\!q}
	\!\leq\!
	|1\!-\!\zeta_1 \overline{w}|
	\exp\Big(\!\big(\!(\zeta_1 \overline{w})^2 - 2\zeta_1 \overline{w}\!\big)\log\Big(\!\frac{N_I}{q}\!\Big)\!\Big)\,,
	\end{equation}
	then
	\begin{equation}\label{bound1}
	\eta^2(\!\gamma_1^\star \, \partial f_1(\!\bm \theta^\star\!)\!)
	\!\leq\!
	2 v_{\zeta_1} \log \Big(\!\frac{N_I}{q}\!\Big)	\!+\! s \!+\! |K_{\zeta_1}^{\neq}| \!+\! \frac{1}{2}|K_{\zeta_1}^=| \!+\!	\frac{4}{5}q\,.
	\end{equation}
	
	\begin{equation*}
	\textrm{If~}
	\frac{q-s}{N_I-q}
	\geq
	|1-\zeta_1 \overline{w}|
	\exp\Big(4\frac{(\zeta_1 \overline{w} -2)\zeta_1 \overline{w}}{|1-\zeta_1 \overline{w}|^2}
	\log\Big(\frac{q}{s}\Big)\Big)\,,
	\end{equation*}
	then
	\begin{equation*}
	\eta^2(\!\gamma_1^\star \, \partial f_1(\!\bm \theta^\star\!)\!)
	\!\leq\!
	\frac{2v_{\zeta_1}}{(1 \!-\! {\zeta_1} \overline{\vw})^2} \log\Big(\!\frac{q}{s}\!\Big)
	\!+\!
    |K_{\zeta_1}^{\neq}| \!+\! \frac{1}{2}|K_{\zeta_1}^=|
	\!+\!
	\frac{9}{5}s\,.	
	\end{equation*}

\end{lemma}

\begin{lemma}
\label{lemma4}
	Let $\bm \theta^\star \in \R^{N_I}$ be an $s$-sparse vector and $\vw \in \R^{N_I}$ be its similar information. Let $f_2$ denote the convex function $ f_2(\bm \theta):=\norm{\bm \theta}_1+\zeta_2/2 \norm{\bm \theta -\vw}_2^2$. 


	\begin{equation}\label{condition2}
	\textrm{If~}
		\frac{q\!-\!s}{N_I\!-\!q}
		\!\leq\!
		|1\!-\!\zeta_2 \overline{w}|
		\exp\Big(\!\big(\!(\zeta_2 \overline{w})^2 - 2\zeta_2 \overline{w}\!\big)\log\Big(\!\frac{N_I}{q}\!\Big)\!\Big)\,,
	\end{equation}
	then
	\begin{equation}\label{bound2}
		\eta^2(\!\gamma_2^\star \, \partial f_2(\!\bm \theta^\star\!)\!)
		\!\leq\!
		2 v_{\zeta_2} \log\Big(\!\frac{N_I}{q}\!\Big)
		\!+\!
		s
		\!+\!
		|K_{\zeta_2}^{\neq}| \!+\! \frac{1}{2}|K_{\zeta_2}^=|
		\!+\!
		\frac{4}{5}q\,.
	\end{equation}
	
	\begin{equation*}
	\textrm{If~}
		\frac{q-s}{N_I-q}
		\geq
		|1-\zeta_2 \overline{w}|
		\exp\Big(4\frac{(\zeta_2 \overline{w} -2)\zeta_2 \overline{w}}{|1-\zeta_2 \overline{w}|^2}
		\log\Big(\frac{q}{s}\Big)\Big)\,,
	\end{equation*}
	then
	\begin{equation*}
		\eta^2(\!\gamma_2^\star \, \partial f_2(\!\bm \theta^\star\!)\!)
		\leq
		\frac{2v_{{\zeta_2}}}{(1 \!-\! {\zeta_2} \overline{\vw})^2} \log\Big(\!\frac{q}{s}\!\Big)
		\!+\!
		|K_{\zeta_2}^{\neq}| \!+\! \frac{1}{2}|K_{\zeta_2}^=|
		\!+\!
		\frac{9}{5}s\,.	
	\end{equation*}

\end{lemma}

Once Assumption~\ref{assumption_l1} holds, the conditions  \eqref{condition1} and \eqref{condition2} hold in this case. 
Combining Lemma~\ref{lemma2}, we have
\begin{align}
  &~ \xi(\mathcal{C}_{f_1} \cap \S^{N_I-1})   
  \leq~ 2 \sqrt{2 v_{\zeta_1} \log \Big(\frac{N_I}{q}\Big)	+s+|K_{\zeta_1}^{\neq}| + \frac{1}{2}|K_{\zeta_1}^=|	+	\frac{4}{5}q}  + 1 \notag \\
  &~ \xi(\mathcal{C}_{f_2} \cap \S^{N_I-1})  
  \leq~ 2 \sqrt{2 v_{\zeta_2} \log \Big(\frac{N_I}{q}\Big)	+s+|K_{\zeta_2}^{\neq}| + \frac{1}{2}|K_{\zeta_2}^=|	+	\frac{4}{5}q} + 1. \notag
\end{align}
When $N_I \gg q$, i.e., the neural network is very sparse, we have $s$, $|K_{\zeta_2}^{\neq}|$, $\frac{1}{2}|K_{\zeta_2}^=|$ and $q$ are small and hence we have 
\begin{align}
  \xi(\mathcal{C}_{f_1} \cap \S^{N_I-1}) &\le  \mathcal{O} \left( \sqrt{v_{\zeta_1} \log ({N_I})	}  \right ), \notag \\
  \xi(\mathcal{C}_{f_2} \cap \S^{N_I-1}) &\le  \mathcal{O} \left( \sqrt{v_{\zeta_2} \log ({N_I})	}  \right ). \notag
\end{align}

In addition, if $\eta \leq \frac{\hat\eta}{\beta R}$, $\bm w = \bm w^{\star}$ and $\gamma \leq 1 / \alpha_f$ we further have for some $\zeta_1>0$

\begin{align}
v_{{\zeta_1} } &= \sum_{n \in I}(\text{sign}(\theta_n^\star)-{\zeta_1} w_n^\star)^2+\sum_{n \in K_{\zeta_1}^{\neq}} ({\zeta_1} |w_n^\star|-1)^2  \notag \\
&\overset{(a)}{\approx} \sum_{n \in I}({\zeta_1} \theta_n^\star - {\zeta_1} w_n^\star)^2 
\overset{(b)}{\leq} {2 {\zeta_1}^2 } \left( \frac{2\gamma_w^2 d_s^2}{\lambda^2} +  \frac{1}{\epsilon^4}  \right)     ,
\end{align}
where $(a)$ is due to Assumptions~\ref{assumption_sign}, and $(b)$ is due to Theorem~\ref{theorem_Base} and Lemma~\ref{lemma8}:

\begin{lemma} \label{lemma8}
Let Assumptions~\ref{assumption_cs} holds. We have 
$$\frac{1}{N} \sum_{i=1}^N \mathbb{E} \left[  \|  \hat{\bm \theta}_i (\bm w^{\star})   -  \bm w^{\star} \|^2_2  \right]   
     \leq  \frac{2\gamma_w^2 d_s^2}{\lambda^2}. $$
\end{lemma}

Meanwhile, if $\eta \leq \frac{\hat\eta}{\beta R}$,  $\bm w = \bm w^{\star}$ and $\gamma \leq 1 / \alpha_f$ we have for some $\zeta_2>0$
\begin{align}
v_{\zeta_2} = &~ \sum_{n \in I}(\text{sign}(\theta_n^\star)+\zeta_2 (\theta_n^\star-w_n^\star))^2+\sum_{n \in K_{\zeta_2}^{\neq}} (\zeta_2 |w_n^\star|-1)^2,  \notag \\
\overset{(a)}{\approx}&~ \sum_{n \in I}(\text{sign}(\theta_n^\star)  + {\zeta_2}( \theta_n^\star -  w_n^\star) )^2  \notag\\
\overset{(b)}{\leq}&~ 2 |I| + {2 {\zeta_2}^2 } \left( \frac{2\gamma_w^2 d_s^2}{\lambda^2} +  \frac{1}{\epsilon^4}  \right)   ,
\end{align}
where $(a)$ is due to Assumptions~\ref{assumption_sign}, and $(b)$ is due to Theorem~\ref{theorem_Base} and Lemma~\ref{lemma8}.
Then we complete the proof.
\end{proof}


\subsection{Proof of Proposition~\ref{proposition_convex_analyse}}


\begin{proof}
{\color{black}
We first prove some interesting properties of $\phi_{\rho}(\bm x)$, which is convex smooth approximation to $\|\bm x\|_1$ and
\begin{align}
    0\leq\nabla^2\phi_{\rho}(x)=\frac{1}{\rho}\left(1-\left(\tanh(x/\rho)\right)^2\right)\leq\frac{1}{\rho}\notag,
\end{align}
which yields $\mu_\phi\left(x - \bar x\right) \leq \nabla\phi_{\rho}(x)-\nabla\phi_{\rho}(\bar x) \leq \frac{1}{\rho}\left(x - \bar x\right)$, where $\mu_\phi=\frac{1}{\rho}\left(1-\left(\tanh(x/\rho)\right)^2\right) \geq 0$, then we have

\begin{align}
     &\left\|\nabla\phi_{\rho}(\bm x)-\nabla\phi_{\rho}(\bm{\bar x})\right\|_2 
     \leq
     \frac{1}{\rho}\left\|\bm x-\bm{\bar x}\right\|_2, \notag
     \\
     &\left\langle \nabla \phi_\rho(\bm x) - \nabla \phi_\rho(\bm{\bar x}) , \bm x-\bm{\bar x} \right\rangle 
     \geq 
     \mu_\phi \left\|\bm x-\bm{\bar x}\right\|_2^2. \notag
\end{align}
Hence, $\phi_{\rho}$ is $\mu_\phi$-strong convex function with $\frac{1}{\rho}$-Lipschitz $\nabla\phi_\rho$. 

We first prove the smooth of $F_i(\bm w)$. We defined $\bar\mu$ as
$$
\bar\mu := \left\{ \begin{array}{l}
\mu_\phi + \mu, ~\text{\rm{if Assumption~1 (a)~holds;}}\\
\mu_\phi - L, ~\text{\rm{if Assumption~1 (b)~holds, and~ $\mu_\phi \geq L$.}}
\end{array} \right.
$$
Then, $\ell_i(\bm \theta_i) + \gamma \phi_\rho(\bm \theta_i)$ is $\bar\mu$-strongly convex with $\gamma \geq 0$. Let $\bar\ell_i(\bm \theta_i) := \frac{1}{\lambda}\big( \ell_i(\bm \theta_i) + \gamma \phi_\rho(\bm \theta_i) \big)$. Since $\bar\ell_i(\bm \theta_i)$ is $\frac{\bar\mu}{\lambda}$-strongly convex, it follows that its conjugate function $\max_{\bm \theta_i \in \mathcal{R}^d} \left\{ \langle \bm \theta_i, \bm w \rangle - \bar\ell_i(\bm \theta_i) \right\}$ is $\frac{\lambda}{\bar\mu}$-smooth. Then, $\bar\ell^*_i(\bm w)$  is $\frac{\lambda}{\bar\mu}$-smooth for
\begin{align}
    \bar\ell^*_i(\bm w) 
    &= \min_{\bm \theta_i \in \mathcal{R}^d} \left\{ \bar\ell_i(\bm \theta_i) - \langle \bm \theta_i, \bm w \rangle\right\}\notag.
\end{align}
 Let $\Lambda^*_i(\bm w) := \lambda \bar\ell^*_i(\bm w) + \gamma_w\phi_\rho(\bm w)$,  then for $\forall \bm w , \bm w^\prime \in \mathbb{R}^d$, we have
\begin{align}
    \| \nabla \Lambda^*_i(\bm w) - \nabla \Lambda^*_i(\bm w^\prime)  \|_2 
    \leq&~ (\frac{\lambda^2}{\bar\mu} + \frac{\gamma_w}{\rho}) \| \bm w - \bm w^\prime \|_2.\notag
\end{align}
Thus $\Lambda^*_i(\bm w)$ is $(\frac{\lambda^2}{\bar\mu} + \frac{\gamma_w}{\rho})$-smooth. Similarly, $F_i(\bm w) = \Lambda^*_i(\bm w) + \frac{\lambda}{2} \| \bm w \|_2^2$ is $L_F$-smooth on $\mathbb{R}^d$ with $L_F = \frac{\lambda^2}{\bar\mu} + \frac{\gamma_w}{\rho} + \lambda$. 

We next prove the strong convexity of $F_i(\bm w)$.  According to the properties of the conjugate function, $\bar\ell^*_i(\bm w)$ is convex. Then, $\Lambda^*_i(\bm w) := \lambda \bar\ell^*_i(\bm w) + \gamma_w\phi_\rho(\bm w)$ is $\gamma_w \mu_\phi$-strong convex. We have $F_i(\bm w) = \Lambda^*_i(\bm w) + \frac{\lambda}{2} \| \bm w \|_2^2$ is $\mu_F$-strong convex on $\mathbb{R}^d$ with $\mu_F = \gamma_w \mu_\phi + \lambda$. We complete the proof.
}
\end{proof}

\subsection{Proof of Lemma~\ref{lemma_theta-theta}}
\begin{proof}
For $\ell_i(\bm \theta_i)$ is $\mu$-strongly convex, then $H_i( {\bm \theta}_i; \bm w_i^{t,r})$ is $\mu$-strongly convex with its unique solution $\hat{\bm \theta}_i (\bm w_i^{t,r})$. Then, we have
\begin{align}
    &~ \| \tilde{\bm \theta}_i (\bm w_i^{t,r})  -  \hat{\bm \theta}_i (\bm w_i^{t,r}) \|_2^2   
    \overset{(a)}{\leq}
    \frac{1}{\mu_F^2}  \|  \nabla H_i( \tilde{\bm \theta}_i; \bm w_i^{t,r}) \|^2_2   \notag\\
    \overset{(b)}{\leq}
    &~ 
    \frac{2}{\mu_F^2} \left( \frac{1}{|\mathcal{D}|^2} \| \sum_{Z_i \in \mathcal{D}_i} \nabla \tilde{\ell}_i(\tilde{\bm \theta}_i;  Z_i)  -  \nabla {\ell}_i(\tilde{\bm \theta}_i) \|^2_2  + v \right) ,   \notag
\end{align}
where $(a)$ is due to $\nabla H_i( \hat{\bm \theta}_i; \bm w_i^{t,r}) = 0$ and $(b)$ is due to Jensen's inequality. Then, we have
\begin{align}
     \mathbb{E}\left[ \| \tilde{\bm \theta}_i(\bm w_i^{t,r}) - \hat{\bm \theta}_i(\bm w_i^{t,r}) \|_2^2  \right]  
    \overset{(a)}{\leq}&~  \frac{2}{\mu_F^2} \left( \frac{1}{|\mathcal{D}|^2} \!\sum_{Z_i \in \mathcal{D}_i} \mathbb{E}_{Z_i} \left[\|  \nabla \tilde{\ell}_i(\tilde{\bm \theta}_i;  Z_i)  \!-\!  \nabla {\ell}_i(\tilde{\bm \theta}_i) \|^2_2 \right]  \!+\! v \right)    \notag\\
    \overset{(c)}{\leq}&~ \frac{2}{\mu^2} \left( \frac{\gamma^2_\ell}{|D|} + v \right) , \notag
\end{align}
where $(a)$ is due to $\mathbb{E}\left[ \|\sum_{i=1}^{M} X_i - \mathbb{E}[X_i] \|^2_2 \right] = \sum_{i=1}^{M} \mathbb{E}\left[ \|X_i - \mathbb{E}[X_i] \|_2 \right]^2 $ with $M$ independent random variables $X_i$ and o Jensen's inequality and $(c)$ is due to Assumption~\ref{assumption_bv}. We finish the proof.
\end{proof}

\subsection{Proof of Lemma~\ref{lemma_g-F}}
\begin{proof}

To facilitate the analysis, we define some additional notations and rewrite the local update. 
\begin{align}
    \bm w_{i}^{t,r+1}=\bm w_{i}^{t,r}-\eta \underbrace{\left[ \lambda\left(\bm w_{i}^{t,r}-\tilde{\bm \theta}_{i}(\bm w_{i}^{t,r})\right)+\gamma_w\nabla\phi_\rho(\bm w_{i}^{t,r})\right]}_{\triangleq  g_{i}^{t,r}},\notag
\end{align}
where $g_{i}^{t,r}$ can be interpreted as the biased estimate of $\nabla F_{i}(\bm w^{t,r}_i)$. We next rewrite the global update as
\begin{align}
    \bm w^{t+1} 
    &=\bm w^{t}-\underbrace{\eta \beta R}_{\triangleq  \tilde{\eta}} \underbrace{\frac{1}{S R} \sum_{i \in \mathcal{S}^{t}} \sum_{r=0}^{R-1} g_{i}^{t,r}}_{\triangleq  g^{t}},\notag
\end{align}
where $\tilde{\eta}$ and $g^{t}$ can be respectively considered as the step size and approximate stochastic gradient of the global update. Then, We have
\begin{align}
    \mathbb{E}\left[  \left\|g_{i}^{t,r}-\nabla F_{i} \left( \bm w^{t} \right) \right\|_2^{2}  \right]   
    \overset{(a)}{\leq}&~    2 \mathbb{E}\left[  \left\|g_{i}^{t,r} \!-\! \nabla F_{i} \left( \bm w_i^{t,r} \right) \right\|_2^{2}   \!+\!    \left\|\nabla F_{i} \left( \bm w_i^{t,r} \right) \!-\! \nabla F_{i} \left( \bm w^{t} \right) \right\|_2^{2} \right]   \notag\\
    \overset{(b)}{\leq}&~ 2 \left(  \lambda^2 \delta^2    +    L_F^2 \| \bm w_i^{t,r} - \bm w^t \|^2_2 \right) ,   \label{eq_g-fi}
\end{align}
where $(a)$ is due to Jensen's inequality, $(b)$ is due to $F_i$ is $L_F$-smooth and Lemma~\ref{lemma_theta-theta}. Then, we bound the drift of the local model form the global model as the following

\begin{align}
     \mathbb{E}\left[ \| \bm w_i^{t,r} - \bm w^t \|^2_2  \right]  =&~ \mathbb{E}\left[  \| \bm w_i^{t,r-1} - \bm w^t -\eta g_i^{t,r-1} \|^2_2 \right]  \notag\\
    \overset{(a)}{\leq}&~ 
    (1 + \frac{1}{R})\mathbb{E}\left[  \| \bm w_i^{t,r-1} - \bm w^t \|^2_2 \right]    +    (1 + R) \eta^2 \mathbb{E}\left[  \| g_i^{t,r-1} \|^2_2 \right]  \notag\\
    \leq&~ 
    \left(1 + \frac{1}{R}  +  4(1 + R) \eta^2 L_F^2 \right)\mathbb{E}\left[  \| \bm w_i^{t,r-1} \!-\! \bm w^t \|^2_2 \right]  \notag\\
    &~ +   2(1 + R) \eta^2\mathbb{E}\left[  \| \nabla F_i(\bm w^{t})\|^2_2  \right] + 4(1 + R) \eta^2\lambda^2 \delta^2,   \notag
\end{align}
where $(a)$ is due to the AM–GM inequality and Jensen's inequality. Let $\eta \leq \frac{1}{2L_F\sqrt{R(1+R)}}  \Leftrightarrow \frac{\beta\sqrt{R}}{2L_F\sqrt{1+R}}   $, we have
$$ 1 + \frac{1}{R}  +  4(1 + R) \eta^2 L_F^2 \leq 1 + \frac{2}{R} $$
$$ 2(1 + R) \eta^2 = \frac{2(1+R)\tilde\eta\eta}{\beta R} \leq  \frac{\tilde\eta  \sqrt{1+R}}{\beta R L_F\sqrt{R}} \leq \frac{2\tilde\eta}{\beta L_F R} $$
$$ 4(1 + R) \eta^2 \leq \frac{4(1+R)}{4L_F^2 R(1+R)} = \frac{1}{L_F^2 R}.   $$
Then,
\begin{align}
    \mathbb{E}\left[ \| \bm w_i^{t,r} \!-\! \bm w^t \|^2_2  \right] 
    \overset{(a)}{\leq}& 
    \frac{1}{L_F^2} \left(\!  \frac{8\tilde\eta L_F}{\beta}  \mathbb{E}\left[  \| \nabla F_i(\bm w^{t})\|^2_2  \right] \!+\!     4\lambda^2 \delta^2   \! \right) , \label{eq_wi-w}
\end{align}
where  $(a)$ is due to unrolling the recurrence formula with $\bm w_i^{t,0} = \bm w^t$ and using $\frac{(1+2/R)^R-1}{2/R} \leq \frac{e^2-1}{2/R} < 4R$. 

By substituting \eqref{eq_wi-w} into \eqref{eq_g-fi}, we have
\begin{align}
    \mathbb{E}\!\left[\!  \left\|g_{i}^{t,r} \!-\! \nabla F_{i} \!\left( \bm w^{t} \!\right) \right\|_2^{2}  \!\right]
    \!\leq\!&   \frac{16\tilde\eta L_F}{\beta} \mathbb{E} \!\left[\! \left\|\nabla F_{i}\left(\!\bm w^{t}\!\right)\right\|_2^{2}\!\right]  \!+\! 10 \lambda^{2} \delta^{2}. \notag
\end{align}
By taking average over $N$ and $R$, we have
\begin{align}
    \frac{1}{N R} \sum_{i, r=1}^{N, R} \mathbb{E}\left[\left\|g_{i}^{t,r}-\nabla F_{i}\left(\bm w^{t}\right)\right\|_2^{2}\right] 
    \overset{(a)}{\leq}&~ \frac{32\tilde\eta L_F}{\beta N}  \sum_{i=1}^{N} \left( \mathbb{E}\left[\left\|\nabla F_{i}\left(\bm w^{t}\right)   -   \nabla F_{i}\left(\bm w^{\star}\right)  \right\|_2^{2}\right] \right) +  8\sigma_{F}^{2}  \notag\\
    &~ +  \left(  2  +  \frac{64\tilde\eta^2 L_F^2 }{\beta^2} \right)\lambda^{2} \delta^{2} \notag\\
    \overset{(b)}{=}& 64\tilde\eta L_F^2  \mathbb{E} \left[     F\left(\bm w^{t}\right)  \!-\!  F\left(\bm w^{\star}\right)     \right]  \!+\!  8\sigma_{F}^{2}  \!+\! \left(  2  \!+\!  64\tilde\eta^2 L_F^2 \right)\lambda^{2} \delta^{2}. 
\end{align}
where $(a)$ is due to Jensen's inequality with setting $\sigma_{F}^{2} \triangleq   \frac{1}{N} \sum_{i}^{N}\left\|\nabla F_{i}\left(\bm w^{\star}  \right)  \right\|_2^{2}$, and $(b)$ is due to $\frac{1}{N}\sum_{i=1}^{N} \nabla F_{i}\left(\bm w^{\star}\right) = \nabla F\left(\bm w^{\star}\right)=0$ with $\beta \geq 1$.

\end{proof}

\subsection{Proof of Lemma~\ref{lemma_Fi-F}}
\begin{proof}
Using the conclusion of Lemma 2 in \cite{t2020personalized}, if Assumption~\ref{assumption_cs} holds, then we obtain Lemma~\ref{lemma_Fi-F}.
\end{proof}

\subsection{Proof of Lemma~\ref{lemma_Fi-F_s}}
\begin{proof}
Using the conclusion of Lemma 4 in \cite{t2020personalized}, if Assumption~\ref{assumption_cs} holds, then we obtain Lemma~\ref{lemma_Fi-F_s}.
\end{proof}

\subsection{Proof of Lemma~\ref{lemma_one_step}}
\begin{proof}
First, we have
\begin{align}
\label{Ewt}
 {\mathbb{E}}\left[\left\|\bm w^{t+1}-\bm w^{\star}\right\|^{2}\right] 
=&~ {\mathbb{E}} \left[\left\|\bm w^{t}-\bm w^{\star}\right\|_2^{2}\right]-2 \tilde{\eta} {\mathbb{E}}\left[\left\langle g^{t}, \bm w^{t}-\bm w^{\star}\right\rangle\right]+\tilde{\eta}^{2} \mathbb{E}\left[\left\|g^{t}\right\|_2^{2}\right]
\end{align}
and the second term of \eqref{Ewt} is as follow
\begin{align}
 -{\mathbb{E}}\left\langle g^{t}, \bm w^{T}-\bm w^{\star}\right\rangle   
=&~ -\frac{1}{N R} \sum_{i, r}^{N, R}(\left\langle g_{i}^{t,r}-\nabla F_{i}\left(\bm w^{t}\right), \bm w^{t}-\bm w^{\star}\right\rangle  
 +   \left\langle\nabla F_{i}\left(w^{t}\right), \bm w^{t}-\bm w^{\star}\right\rangle)\notag\\
\overset{(a)}{\leq}&~ 
\frac{1}{2N R} \sum_{i, r}^{N, R}\left(\!\frac{\left\|g_{i}^{t,r}\!-\!\nabla F_{i}\left(\bm w^{t}\right)\right\|_2^{2}}{\mu_{F}}\!\right)\!+\!F\left(\!\bm w^{\star}\!\right)\!-\!F\left(\!\bm w^{t}\!\right).\label{eq-la3-2}
\end{align}
where $(a)$ follows by the AM–GM inequality and $F_i$ is $\mu_F$-strongly convex.

From equations $(18)$ and $(19)$ in \cite{t2020personalized}, we have
\begin{align}
    \left\|g^{t}\right\|^{2} \leq& \frac{3}{N R} \sum_{i, r}^{N, R}\left\|g_{i}^{t,r}-\nabla F_{i}\left(\bm w^{t}\right)\right\|_2^{2}  
     +   3 \left\|\frac{1}{S} \sum_{i \in \mathcal{S}^{t}} \nabla F_{i}\left(\bm w^{t}\right)-\nabla F\left(\bm w^{t}\right)\right\|_2^{2}    
     + 6 L_{F}\left(F\left(\bm w^{t}\right)-F\left(\bm w^{\star}\right)\right).\label{eq-la3-3}
\end{align}

By substituting \eqref{eq-la3-2}, \eqref{eq-la3-3} into \eqref{Ewt}, we finish the proof of Lemma~\ref{lemma_one_step}.
\end{proof}

\subsection{Proof of Lemma~\ref{lemma2}}

\begin{proof} For any vector $\vd \in \mathcal{C}_f$, we have
	\begin{equation} \label{neq: convex_cone3}
	 \ip{\vd}{\gamma\vu} \le  0,
	\end{equation}
where $\vu \in \partial f(\bm \theta^\star)$. Then we obtain
\begin{align} \label{neq: convex_cone4}
	\ip{\vd}{\vg} &\le  \ip{\vd}{\vg-\gamma \vu} \le \norm{\vd}_2 \norm{\vg-\gamma  \vu}_2,
\end{align}
where $\vg \sim \mathcal{N}(0,\vI_{N_I})$. Choosing $\vu$ such that $\norm{\vg-\gamma \vu}_2=\dist(\vg, \gamma \partial f(\bm \theta^\star))$, we get
\begin{align} \label{neq: convex_cone5}
	\ip{\vd}{\vg} \le  \norm{\vd}_2 \norm{\vg-\gamma  \vu}_2 = \norm{\vd}_2 \dist(\vg, \gamma \, \partial f(\bm \theta^\star)),
\end{align}
where $\dist(\vg, \gamma \partial f(\bm \theta^\star))=\inf_{\vu \in\partial f(\bm \theta^\star)} \norm{\vg-\gamma \vu}_2$.

Taking expectation over $\vg$ yields
\begin{align}
	w(\mathcal{C}_f \cap \S^{N_I-1}) = \E \sup_{\vd } \ip{\vd}{\vg} \le& \E \dist(\vg, \gamma \, \partial f(\bm \theta^\star))   \notag\\
	\le& \eta(\gamma \, \partial f(\bm \theta^\star)),
\end{align}
where $\vd \in \mathcal{C}_f \cap \S^{N_I-1}$, the first inequality uses $\norm{\vd}=1$ for $\vd \in \S^{N_I-1}$ and the second inequality applies Jenson's inequality.
Using the relationship between Gaussian width and Gaussian Complexity  \eqref{Relation} completes the proof.
\end{proof}


\subsection{Proof of Lemma~\ref{lemma3}}

\begin{proof}
	Using the definition of Gaussian squared distance, we obtain
	\begin{align*}
	\eta^2(\gamma_1^\star \, \partial f_1(\bm \theta^\star))=&~ \min_{\gamma \ge 0}  \E \left[ \dist^2(\vg, \gamma \cdot (\partial \norm{\bm \theta^\star}_1 - \zeta_1 \vw))\right]\\
	\le&~ \underbrace{\sum_{i \in I} \E_{\vg_i}\[\dist^2(\vg_i,\gamma \text{sign}(\bm \theta_i^\star)-\gamma\zeta_1 \vw_i)\]}_{\Gamma_1}  
	 +
	\underbrace{\sum_{i \in I^c} \E_{\vg_i}\[\dist^2(\vg_i,I (-\gamma\zeta_1 \vw_i ,\gamma))\]}_{\Gamma_2},
	\end{align*}
	where $I(x,y)=[x-y,x+y]$, for $y\ge0$.
	
	To bound $\Gamma_1$, we have
	\begin{align*}
	\Gamma_1 &= \sum_{i \in I} \E_{\vg_i}\left[\left(\vg_i-(\gamma \text{sign}(\bm \theta_i^\star)-\gamma\zeta_1 \vw_i)\right)^2\right] 
	=s+ \gamma^2\left[\sum_{i \in I}(\text{sign}(\bm \theta_i^\star)-\zeta_1 \vw_i)^2\right].
	\end{align*}
	
	To bound $\Gamma_2$, it is exactly the same as (91b) of  [27]. So we cite the result as follows
	\begin{align*}
	\Gamma_2 \le&~ \gamma^2\left[ \sum_{i \in K_{\zeta_1}^{\neq}} (\zeta_1 |\vw_i|-1)^2\right] + |K_{\zeta_1}^{\neq}| + \frac{1}{2}|K_{\zeta_1}^=| 
	 +  2|I^cJ| \frac{\varphi(\gamma(1-\zeta_1 \overline{w}))}{\gamma|1-\zeta_1 \overline{w}|}  + 2 |I^cJ^c| \frac{\varphi(\gamma)}{\gamma}.
	\end{align*}

	Combining $\Gamma_1$ and $\Gamma_2$ yields
	\begin{align*}
	\eta^2(\gamma_1^\star \, \partial f_1(\bm \theta^\star))
	\le&~ \Gamma_1+\Gamma_2  
	\le~  s+  \gamma^2  v_{{\zeta_1}}  + |K_{\zeta_1}^{\neq}| + \frac{1}{2} |K_{\zeta_1}^=| 
	 + 2|I^cJ| \frac{\varphi(\gamma(1-\zeta_1 \overline{w}))}{\gamma|1-\zeta_1 \overline{w}|} + 2 |I^cJ^c| \frac{\varphi(\gamma)}{\gamma},
	\end{align*}
	where
	$v_{{\zeta_1}} = \sum_{i \in I}(\text{sign}(\bm \theta_i^\star)-{\zeta_1} \vw_i)^2+\sum_{i \in K_{\zeta_1}^{\neq}} ({\zeta_1} |\vw_i|-1)^2$.
	The following deviation is almost the same as (97) of \cite{mota2017compressed}  by replacing $v_{\b}$ by $v_{\zeta_1}$. Thus we get the desired results.
\end{proof}

\subsection{Proof of Lemma~\ref{lemma4}}
\begin{proof}
It's straightforward to get the upper bound of $\eta^2(\gamma_2^\star \, \partial f_2(\bm \theta^\star))$ by simply replacing $v_{{\zeta_1}}$ with $v_{\zeta_2}$, see \cite[Theorem 14]{mota2017compressed} for details.
\end{proof}

\subsection{Proof of Lemma~\ref{lemma8}}

\begin{proof}
We have
    \begin{align}
        \frac{1}{N}\sum_{i=1}^{N} \E \left[  \| \hat{\bm \theta}_i (\bm w^{\star})   -  \bm w^{\star} \|^2_2  \right]  
        \overset{(a)}{\leq}&~ 
        \frac{2}{N}\sum_{i=1}^{N} \E \left[  \frac{1}{\lambda^2} \| \nabla H_i(\hat{\bm \theta}_i ;\bm w^{\star})  \|^2_2  +  \frac{\gamma_w^2 d_s^2}{\lambda^2} \right] 
        \overset{(b)}{=} \frac{2\gamma_w^2 d_s^2}{\lambda^2}, \notag
    \end{align}
where $(a)$ follows by Jensen's inequality and \eqref{eq_theta-w_2} and $(b)$ is due to $\nabla H_i(\hat{\bm \theta}_i ;\bm w^{\star}) = 0$. Then we complete the proof.
\end{proof}

\end{document}